%% file: main.tex
\documentclass[10pt,twocolumn,letterpaper]{article}

\usepackage[pagenumbers]{cvpr} %

\input{preamble}

\definecolor{cvprblue}{rgb}{0.21,0.49,0.74}
\usepackage[pagebackref,breaklinks,colorlinks,citecolor=cvprblue]{hyperref}
\usepackage[percent]{overpic}
\usepackage{xcolor}
\usepackage{cuted}
\usepackage{multirow}
\usepackage{pifont}

\input{defs.tex}
\title{\acronym: \papertitle}

\author{Yun-Jin Li$^{\star1}$
\and
Mariia Gladkova$^{\star1,2\dagger}$
\and
Yan Xia$^{1,2}$
\and
Daniel Cremers$^{1,2}$\\
{\centering $^1$ TU Munich $^2$ Munich Center for Machine Learning}\\
{\tt \centering \small \{yunjin.li, mariia.gladkova, yan.xia, cremers\}.@tum.de}
}

\begin{document}

\maketitle
\let\oldthefootnote\thefootnote
\renewcommand{\thefootnote}{\fnsymbol{footnote}} 
\footnotetext[2]{Corresponding author. $^\star$ Equal contribution.} 
\let\thefootnote\oldthefootnote

\begin{strip}
\centering
\begin{overpic}[width=1.0\linewidth]{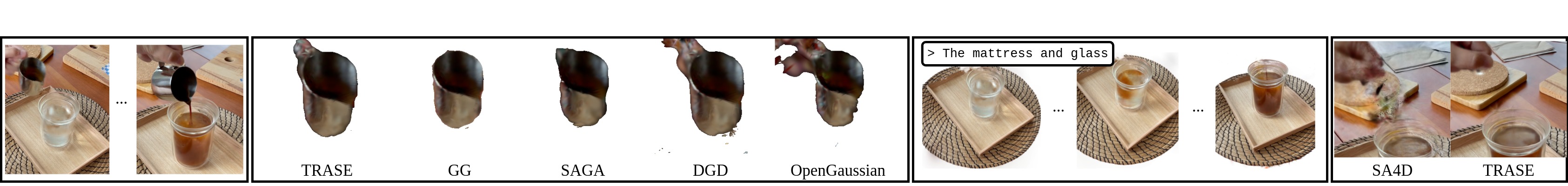}
\put(2, 10){{\scriptsize \shortstack{Rendered Novel Views}}}
\put(29, 10){{\scriptsize \shortstack{Click Prompt Segmentation}}}  
\put(64, 10){{\scriptsize \shortstack{Text Prompt Segmentation}}}   
\put(88, 10){{\scriptsize \shortstack{Object Removal}}}

\end{overpic}
\vspace{-7mm}
\captionof{figure}{We propose TRASE, a novel tracking-free 4D segmentation approach. TRASE achieves superior object segmentation from click prompts and further supports interactive editing tasks such as object removal and text-prompt-based segmentation.}
\label{fig:teaser_with_comprison}
\vspace{-3mm}
\end{strip}

\input{sec/0_abstract}
\vspace{-5mm}
\input{sec/1_intro}
\input{sec/2_related}
\input{sec/3_method}

\input{sec/4_experimental}
\input{sec/5_ablations}
\input{sec/6_downstream_apps}
\input{sec/7_conclusion}

\input{sec/X_suppl}
\clearpage
\clearpage
{
    \small
    \bibliographystyle{ieeenat_fullname}
    \bibliography{main}
}

\end{document}

%% file: preamble.tex
\usepackage[dvipsnames]{xcolor}

%% file: defs.tex
\newcommand{\papertitle}{Tracking-free 4D Segmentation and Editing}
\newcommand{\acronym}{TRASE}

%% file: sec/0_abstract.tex
\begin{abstract}
Understanding dynamic 3D scenes is crucial for extended reality (XR) and autonomous driving. Incorporating semantic information into 3D reconstruction enables holistic scene representations, unlocking immersive and interactive applications. To this end, we introduce \acronym{}, a novel tracking-free 4D segmentation method for dynamic scene understanding. \acronym{} learns a 4D segmentation feature field in a weakly-supervised manner, leveraging a soft-mined contrastive learning objective guided by SAM masks. The resulting feature space is semantically coherent and well-separated, and final object-level segmentation is obtained via unsupervised clustering. This enables fast editing, such as object removal, composition, and style transfer, by directly manipulating the scene's Gaussians. We evaluate \acronym{} on five dynamic benchmarks, demonstrating state-of-the-art segmentation performance from unseen viewpoints and its effectiveness across various interactive editing tasks. Our project page is available at: \href{https://yunjinli.github.io/project-sadg/}{https://yunjinli.github.io/project-sadg/}
\end{abstract}

%% file: sec/1_intro.tex
\section{Introduction}
\label{sec:intro}

\begin{figure*}
  \centering
  \includegraphics[width=1.0\linewidth]{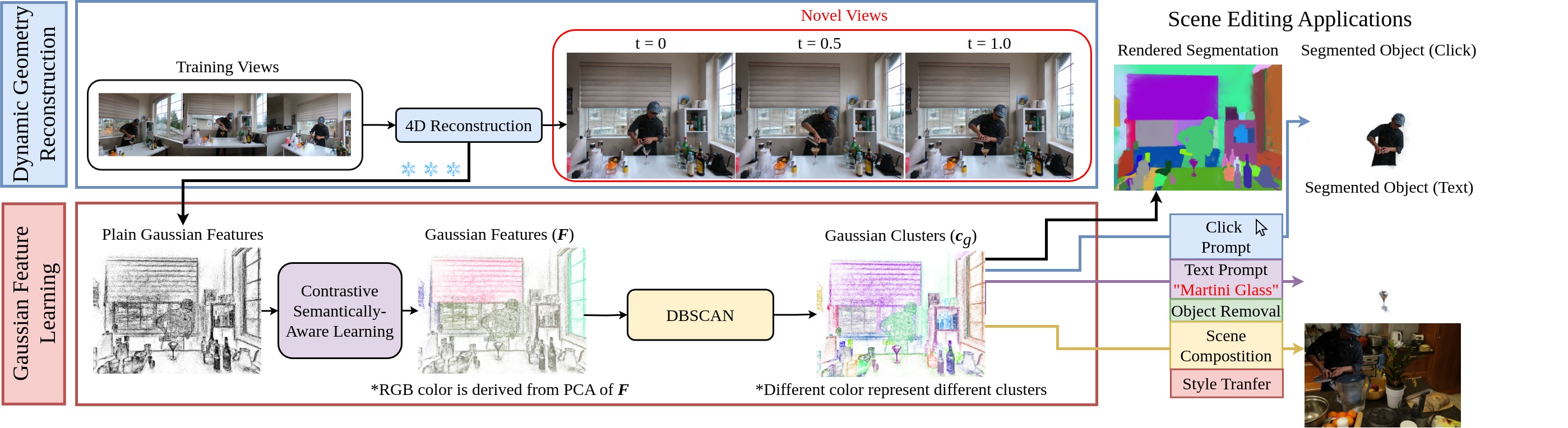}
  \caption{\textbf{Our pipeline.} \acronym{} consists of two main components: dynamic geometry reconstruction (\cref{subsec:georecon}) and Gaussian feature learning (\cref{subsec:gaussfeat}). We adopt the approach from~\cite{yang2024deformable} to effectively learn dynamic 3D reconstruction. 
  Given a 4D reconstruction, we learn Gaussian features $\boldsymbol{F} \in \mathbb{R}^{N \times 32}$ using a novel contrastive learning objective guided by SAM~\cite{kirillov2023sam} masks. Once trained, we apply clustering~\cite{ester1996density} directly to the learned features, enabling segmentation field rendering. Our representation supports various scene-editing applications, including object segmentation via click/text prompts in our GUI, object removal, and scene composition.
  }
  \label{fig:teaser}
  \vspace{-2mm}
\end{figure*}

Humans experience the world in motion, where objects persist over time despite changes in pose, shape, and appearance. Our ability to recognize objects remains stable, even as we move and observe them from different vantage points. This phenomenon suggests the existence of neural 3D representations, which preserve semantic consistency across both space and time. Such representations are central for applications in augmented reality, gaming, and autonomous systems, where interactive manipulation of dynamic scenes requires object-level segmentation that is spatio-temporally consistent and efficient.

Neural Radiance Fields (NeRFs)~\cite{mildenhall2021nerf} have significantly advanced 3D reconstruction by enabling photorealistic novel view synthesis, but their implicit nature makes them computationally too expensive for real-time interaction. Gaussian Splatting (3DGS)~\cite{kerbl3Dgaussians} has recently emerged as a powerful alternative, achieving real-time rendering with explicit and compact scene representations. Dynamic extensions of 3DGS~\cite{luiten2023dynamic, yang2024deformable} allow faithful reconstruction of moving scenes, but they lack built-in object segmentation and rely purely on geometry and color. While several works have integrated semantic information into static 3D Gaussian representations~\cite{cen2023saga, ye2023gaussian}, they fail to handle motion, deformation, and viewpoint variation in dynamic scenes.

Only a handful of works have explored the unification of semantics and dynamics in 3DGS~\cite{ji2024segment, labe2024dgd}.
SA4D~\cite{ji2024segment} incorporates segmentation into dynamic Gaussian splatting, using video object trackers to match masks across views similar to Gaussian Grouping \cite{ye2023gaussian}. However, approaches using these trackers often suffer from identity switches, causing inconsistent segmentation in multi-view settings - particularly in the presence of occlusions, non-rigid deformations, or rapid motion, as shown in~\cref{fig:deva_fail}.
DGD~\cite{labe2024dgd} addresses temporal consistency by distilling semantically-aware features from foundation models like CLIP~\cite{wang2022clip} and DINOv2~\cite{oquab2023dinov2}, enabling tracking. Yet, these features are high-dimensional, leading to significant computational overhead from both the large memory requirements and slow training.

In this work, we present \acronym{}, a tracking-free segmentation framework for dynamic scene understanding.
\acronym{} learns temporally consistent segmentation features in a contrastive setting using only 2D binary masks. 
Unlike prior contrastive segmentation methods~\cite{silva2024contrastive, bhalgat2023contrastive, ying2024omniseg3d}, which explicitly enforce cluster formation during training and promote statistical consistency of features,
\acronym{} applies contrastive learning directly at the pixel-pair level in a fine-grained rendered feature space, deferring object clustering to a final post-training step.
This design enables \acronym{} to learn a segmentation field with both high temporal consistency and sharp, well-defined object boundaries. It also effectively mitigates common artifacts, such as floaters and out-of-FoV errors, that frequently degrade prior contrastive learning approaches.

To the best of our knowledge, we are the first to introduce a benchmark for segmentation in novel views of dynamic scenes. While established datasets like DAVIS~\cite{Perazzi2016, Pont-Tuset_arXiv_2017, Caelles_arXiv_2018, Caelles_arXiv_2019} and VOS~\cite{vos2018, vos2019, vos2022} address dynamic video segmentation, none evaluate segmentation quality in unseen viewpoints. Moreover, our benchmark is significantly larger than the evaluation protocol suggested in~\cite{labe2024dgd}, as our test sequences comprise single- and multi-view scenes and captures real-world scenarios. Beyond achieving state-of-the-art segmentation accuracy, \acronym{} is tailored for real-time interactive scene editing. Our compact 32-dimensional feature space requires minimal post-processing and integrates seamlessly into object manipulation tasks such as style transfer, recoloring, composition, and removal.

Our contributions can be summarized as follows:
\begin{itemize}
\item We propose \acronym{}, a new approach for multi-view consistent segmentation of dynamic scenes without tracking labels. To achieve this, we design a novel contrastive learning objective based on differentiable feature rendering and contrastive learning, ensuring segmentation stability across time and views.
\item We offer a broad benchmark for segmentation in novel views of dynamic scenes, which encompass a broad spectrum of camera setups and captured scenarios. 
\item We extensively evaluate our method on five dynamic datasets and achieve state-of-the-art segmentation performance in both single- and multi-view settings.
\item We demonstrate the applicability of our feature space to several scene editing tasks, including object removal, style transfer, and scene composition. 
\end{itemize}

%% file: sec/2_related.tex
\section{Related Works}
\label{sec:related}
\subsection{3D Segmentation}
\noindent\textbf{In static scenes.} With the emergence of various vision foundation models, such as Segment Anything Model (SAM) \cite{kirillov2023sam}, DINO \cite{caron2021emerging}, and DINOv2 \cite{oquab2023dinov2}, researchers start to integrate the semantics from these foundation models into their reconstruction. SA3D~\cite{cen2023segment} is the pioneering method that extends 2D SAM masks across images into 3D consistent object masks. GARField~\cite{kim2024garfield} introduces hierarchical grouping, leveraging the physical scales of 2D masks to group objects of different sizes and represent scenes at different granularity scales. Gaussian Grouping~\cite{ye2023gaussian} first proposes to utilize consistent object IDs generated by video tracker DEVA~\cite{cheng2023tracking} to render object IDs into camera views. Concurrent work such as SAGA \cite{cen2023saga} utilizes the masks generated from SAM, combined with scales obtained from 3D data inspired by GARField, to segment objects of various sizes. GAGA~\cite{lyu2024gaga} proposes a mask group identity assignment technique using a 3D-aware memory bank to compensate for inaccurate semantic labels and use group IDs as pseudo-labels for identity encoding. Our model implicitly learns multi-view consistent semantic Gaussian features using our designed contrastive loss in a single training stage without identity assignments.

\noindent\textbf{In dynamic scenes.} To the best of our knowledge, only two prior works, SA4D \cite{ji2024segment}, and DGD \cite{labe2024dgd}, address the segmentation task in dynamic 3DGS. DGD extends the rasterization pipeline of 3DGS to be able to render the semantic features of each Gaussian. The rendered Gaussian feature map is then trained with the supervision of DINOv2 \cite{oquab2023dinov2} or CLIP \cite{radford2021learning} to mimic the output from their feature extractors. The drawbacks of such supervision are longer training times and bigger 3DGS models, as the dimensions of the features generated by DINOv2 and CLIP are quite large (384 and 512), and the Gaussian features are also stored. Following the identity encoding concept from Gaussian Grouping and its derivatives \cite{dou2024cosseggaussians, lyu2024gaga}, SA4D introduces a temporal identity feature field. As the method relies on consistent object IDs for supervision, it is sensitive to temporal tracking inaccuracies due to occlusion or fast motion. A concurrent work, Split4D~\cite{hu2025split4d}, follows several proposed design choices, including single-frame supervision and a contrastive learning objective, and utilizes a recent 4D reconstruction model to achieve 4D segmentation. Our \acronym{} does not rely on tracking labels and effectively handles single- and multi-view data with fast rendering times, minimal post-processing, and low memory footprint.
\begin{figure}
  \centering
  \begin{overpic}[width=1.0\linewidth]{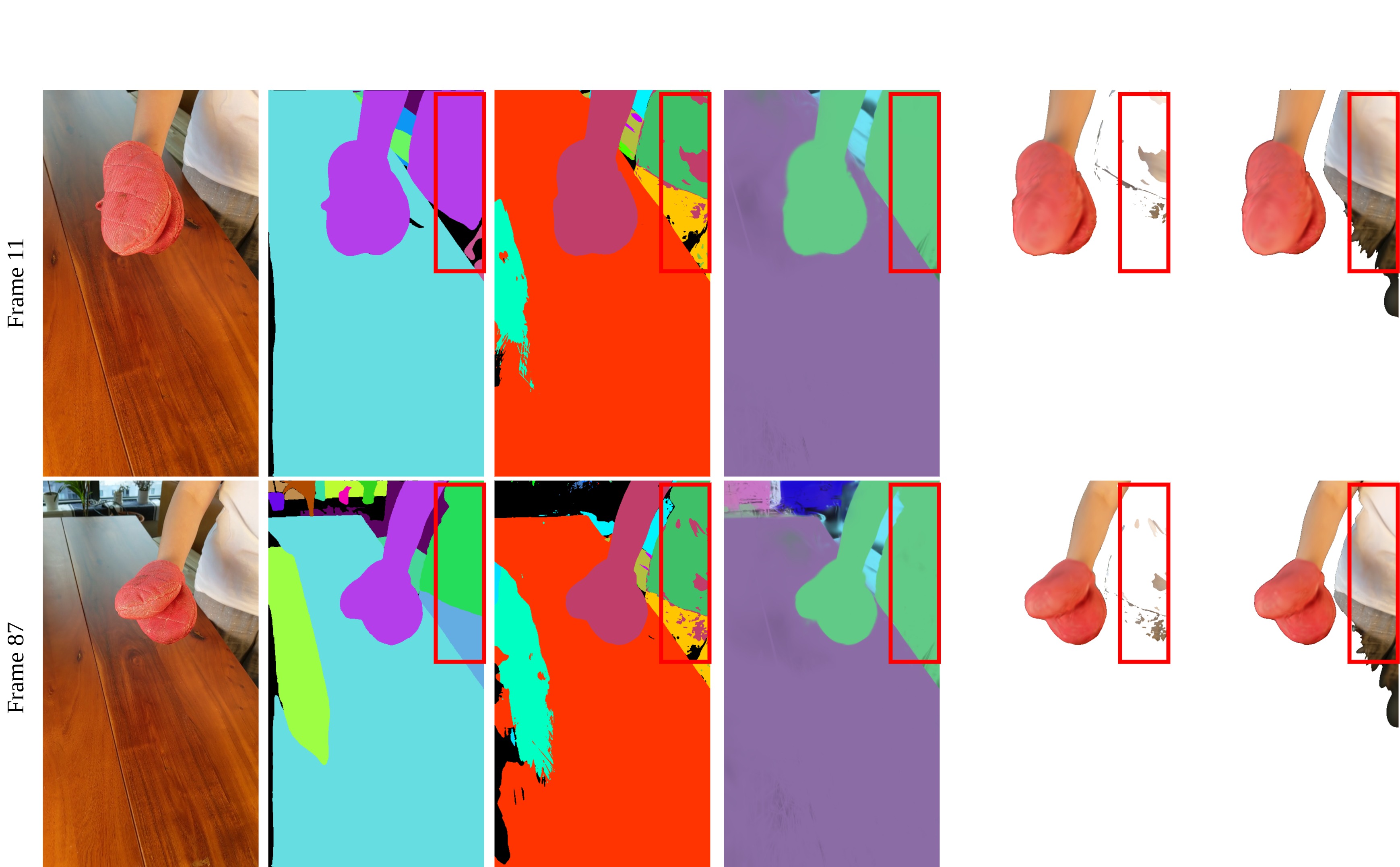}
      \put(8, 58){\fontsize{5}{6}\selectfont {\shortstack{RGB}}}
      \put(18, 58){\fontsize{5}{6}\selectfont {\shortstack{Mask Associatation \\ via Object Tracker \\ (DEVA)}}}
      \put(36, 58){\fontsize{5}{6}\selectfont {\shortstack{Class-agnostic \\ Segmentation via \\ Dynamic GG}}}
      \put(52, 58){\fontsize{5}{6}\selectfont {\shortstack{Class-agnostic \\ Segmentation via \\ \acronym{} (Ours)}}}
      \put(68, 58){\fontsize{5}{6}\selectfont {\shortstack{Segmented Object \\ via \\ Dynamic GG}}}
      \put(88, 58){\fontsize{5}{6}\selectfont {\shortstack{Segmented \\ Object via \\ \acronym{} (Ours)}}}
  \end{overpic}
  \caption{Example of a failure case from the video tracker DEVA \cite{cheng2023tracking}. Different colors refer to various object IDs associated by the model. We can observe that due to the inconsistent presence of the human torso in the video, DEVA fails to provide reliable and consistent object masks for supervision, resulting in a noisy class-agnostic segmentation and poorly segmented objects for dynamic Gaussian Grouping \cite{ye2023gaussian} (GG). We also provide a video of this example in the supplementary materials.
  }
  \label{fig:deva_fail}
  \vspace{-10pt}
\end{figure}
\subsection{Contrastive Learning for 3D Segmentation}
A number of recent works define multi-view 3D segmentation as a contrastive learning objective. Contrastive Gaussian Clustering (CGC)~\cite{silva2024contrastive} training aims to maximize intra-cluster similarity in the 2D rendered feature space by enforcing the feature's proximity with its corresponding cluster's mean feature and ensuring its spatial consistency. OmniSeg3D~\cite{ying2024omniseg3d} follows a similar CGC objective and supervises 3D point clustering using SAM-based image patches. ContrastiveLift~\cite{bhalgat2023contrastive} proposes a slow-fast contrastive fusion strategy, where a slow network aggregates global object-level information while a fast network refines per-point features. OpenGaussian~\cite{wu2025opengaussian} enforces intra-cluster similarity and inter-cluster dissimilarity of rendered features and further trains their discretized version. Split4D~\cite{hu2025split4d} utilizes the same objective as~\cite{silva2024contrastive} and additionally enforces the features to align with DINOv2~\cite{oquab2023dinov2} in the early training stage. Our \acronym{} proposes a contrastive objective based on soft sample mining of pixel pairs, which focuses on only imperfect features and allows us to learn the segmentation field effectively in 4D. No costly distillation is required as we supervise the training with binary 2D masks.
\subsection{Scene Editing Applications}
Learning a semantically aware latent space enables interactive scene editing at the object level. Existing radiance-field editing methods are either restricted to single objects~\cite{guo2020osf, wang2022clip} or rely on per-object scene decompositions~\cite{yang2021learning, kundu2022panoptic}, which limit editing to a fixed set of instances. Other approaches, such as DINO~\cite{oquab2023dinov2}, distill features into volumetric fields~\cite{tschernezki2022neural}, but their implicit representations require expensive re-rendering for consistency, hindering real-time interaction. Gaussian-based methods ~\cite{kerbl3Dgaussians} alleviate NeRF runtime issues via explicit point-based primitives: SAGA~\cite{ye2023gaussian} enables local Gaussian editing, and Feature3DGS~\cite{zhou2024feature} leverages 2D foundation models like SAM~\cite{ravi2024sam2} and LSeg~\cite{li2022lseg} for promptable, language-guided 3D edits, yet all focus on static scenes. In contrast, our method edits dynamic objects with temporally consistent results and supports diverse tasks, including style transfer, object removal, and composition~\cite{labe2024dgd}. While SA4D~\cite{ji2024segment} offers semantic scene editing, it requires prior object labels and limited interaction. In contrast, our framework provides a user-friendly graphical interface that supports simple text prompts and mouse-based selection.

%% file: sec/3_method.tex
\section{Method}
\label{sec:method}
In this section, we introduce our novel \acronym{}. As illustrated in \cref{fig:teaser}, the pipeline comprises two main components: dynamic geometry reconstruction (\cref{subsec:georecon}) and Gaussian feature learning (\cref{subsec:gaussfeat}). We adopt the Deformable-3DGS~\cite{yang2024deformable} pipeline to reconstruct the 4D scene. Once the 4D reconstruction is learned, we freeze this representation and proceed with Gaussian feature learning using SAM~\cite{kirillov2023sam} masks and our novel contrastive learning objective in the rendered feature space. 
\subsection{Dynamic Geometry Reconstruction}
\label{subsec:georecon}

We follow the approach proposed in~\cite{yang2024deformable} and describe it briefly for completeness. Unlike Dynamic3DGS~\cite{luiten2023dynamic} model, where the reconstruction is stored per frame, resulting in high memory consumption, an MLP learns per-Gaussian deformation $(\delta \boldsymbol{x}_i, \delta \boldsymbol{r}_i, \delta \boldsymbol{s}_i)$ with respect to the static canonical space $\mathcal{G}_c$, which is defined as
$\mathcal{G}_c=\{\boldsymbol{x}_i \in \mathbb{R}^3, \boldsymbol{r}_i \in \mathbb{R}^4, \boldsymbol{s}_i \in \mathbb{R}^3, \alpha_i \in \mathbb{R}, \boldsymbol{sh}_i \in \mathbb{R}^{3 \times (D_{max} + 1)^2}\}$ for $i=1, ..., N$.
$\boldsymbol{x}_i$ is the center position of the i-th Gaussian, $\boldsymbol{r}_i$ is the quaternion representing the rotation, $\boldsymbol{s}_i$ is the scaling vector, $\alpha_i$ refers to the opacity, and $\boldsymbol{sh}_i$ is the Spherical Harmonic Coefficients encoding the color information ($D_{max}=3$) as per \cite{yang2024deformable}.

The resulting Gaussians $\mathcal{G}_t$ at timestamp $t$ are defined as
\begin{equation}
    \mathcal{G}_t=\{\boldsymbol{x}_i + \delta \boldsymbol{x}_i, \boldsymbol{r}_i + \delta \boldsymbol{r}_i, \boldsymbol{s}_i + \delta \boldsymbol{s}_i, \alpha_i, \boldsymbol{sh}_i \}_{i=1, ..., N}.
    \label{eq:Gt}
\end{equation}

Once $\mathcal{G}_t$ is obtained, we proceed with the standard rasterization pipeline \cite{kerbl3Dgaussians} to render the image $\boldsymbol{I}_r$. The color loss $\mathcal{L}_\text{color}$ is computed with the ground truth image $\boldsymbol{I}_\text{GT}$ as 
\begin{equation}
\mathcal{L}_\text{color} = (1-\lambda)\mathcal{L}_1 (I_\text{GT}, I_r) +\lambda \mathcal{L}_\text{D-SSIM}(I_\text{GT}, I_r),
\label{eq:L_rendered}
\end{equation}
where $\lambda$ refers to the weighting parameter for the structural similarity loss term.

\subsection{Gaussian Feature Learning}
\label{subsec:gaussfeat}
\textbf{Summary.} Our feature learning augments each 3D Gaussian with a 32-dimensional feature vector, enabling segmentation without explicit tracking labels. It is worth noting that, in contrast to view-dependent spherical encoding, our semantic feature vector is consistent across different views and time. Thanks to differentiable feature rendering~\cite{zhou2024feature} and SAM~\cite{kirillov2023sam} masks, we construct and explicitly enforce pixel-mask associations in the Gaussian feature field.
Our novel contrastive learning objective is designed to be efficient by sub-sampling and effective by integrating SAM~\cite{kirillov2023sam} masks. 
Using only 2D guidance and without any object labels, Gaussians are trained to separate per-pixel rendered features from different segments and align those within the same segment.
To achieve this, we design a soft-mining objective. Unlike hard mining, which selects pairs strictly by similarity, soft mining uses a relaxed, existential rule that increases pixel coverage and makes training less sensitive to noisy feature similarities.
Finally, the learned features are globally clustered, producing multi-view consistent segmentation.
\begin{figure}[htpb]
  \centering
  \includegraphics[trim={0 0 0 0.3cm},clip,width=0.45\textwidth]{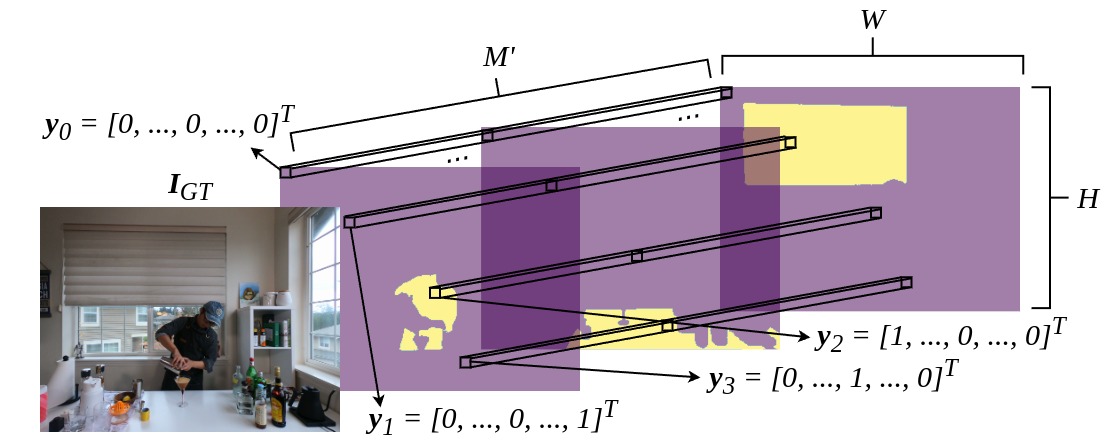}
  \caption{The illustration of pixel-mask correspondence vector $\boldsymbol{y}_{i}$. The pixel-mask correspondence vector is constructed from a subset of SAM masks $\boldsymbol{\mathcal{M}}_{SAM}$ for a given image $\boldsymbol{I}_{GT}$.}
  \vspace{-3pt}
  \label{fig:gt_vec}
\end{figure}

\noindent\textbf{Our notation.}
A generated set of SAM masks $\boldsymbol{\mathcal{M}}_\text{SAM}$ comprise $M'$ binary masks of a given image $\boldsymbol{I}_\text{GT}$, where each mask $\boldsymbol{M}_i$ corresponds to a different segment.
With this, we define the pixel-mask correspondence vector $\boldsymbol{y}_{i}$, which captures information across all binary masks in $\boldsymbol{\mathcal{M}}_\text{SAM}$ for a certain pixel coordinate ($u_i$, $v_i$). This vector is similar to a one-hot encoding technique; however, there is no guarantee that a pixel belongs to a single mask due to 2D segmentation inaccuracies. We visually demonstrate the formulation of $\boldsymbol{y}_{i}$ in \cref{fig:gt_vec}.

The similarities of pixel-mask vectors for a pair of pixels ($i$, $j$) can be retrieved by computing the Gram matrix of their stacked matrix $\boldsymbol{Y} \in \{0, 1\}^{N_p \times M'}$, where $N_p$ refers to the number of sampled pixels. All entries of the resulting Gram matrix are then bounded to either 0 (negative pair) or 1 (positive pair) to form a mask-based correspondence matrix $\boldsymbol{C} \in \{0, 1\}^{N_p \times N_p}$ among sampled pixels, formally defined as
\begin{equation}
    \label{eq:corrmat}
    \boldsymbol{C}(i, j)=
    \begin{cases}
        1,  \quad \text{if} \quad \boldsymbol{y}_{i}^T \cdot \boldsymbol{y}_{j} > 0 \\
        0,  \quad \text{otherwise.}
    \end{cases}
\end{equation}
Analogous to the binary mask-based Gram matrix $\boldsymbol{C}$, we compute a feature-based matrix $\boldsymbol{C}_F$, capturing the pairwise similarities between the rendered features $\{\boldsymbol{I}_F(u_i, v_i)\}_{i=1, ..., N_p}$.
Following common practices~\cite{cen2023saga}, we consider a smooth version of Gaussian features, as it is more robust than considering individual features, which might have high similarity scores corresponding to unrelated objects. The smooth Gaussian feature is computed as the mean feature of its $K$ nearest neighbors~\cite{cover1967nearest} based on the Gaussian's 3D position.

\noindent\textbf{Our learning objective.}
To robustly learn multi-view coherent feature embeddings, we adopt a soft-mined contrastive learning objective. Unlike standard hard mining, which selects positive or negative pairs strictly based on their individual similarity values, soft mining employs an existential criterion: a positive pixel pair ($i$,$j$) is considered into final positive loss computation $\mathcal{L}_\text{pos}$ if there exists any positive pair involving $i$ whose similarity falls below the positive threshold $\tau_p$. Similarly, a pair is included in the negative loss if there exists any negative pair involving $i$ whose similarity exceeds the negative threshold $\tau_n$. By design, the soft-mined strategy broadens pixel coverage during training, in contrast to naive hard sampling that typically concentrates on a sparse set of imperfect pixel pairs and leaves many pairs untrained.
We further restrict the selection to non-redundant pairs by considering only the upper-triangular portion of the matrices ($i < j$). Formally, the soft-mined masks are defined as
\begin{align}
    \boldsymbol{M}^+_{ij} &= \big(\exists k \; : \; \boldsymbol{C}_{ik}=1 \wedge \boldsymbol{C}_{\boldsymbol{F},ik} < \tau_p \big) \;\wedge\; (i < j), \\
    \boldsymbol{M}^-_{ij} &= \big(\exists k \; : \; \boldsymbol{C}_{ik}=0 \wedge \boldsymbol{C}_{\boldsymbol{F},ik} > \tau_n \big) \;\wedge\; (i < j).
\end{align}
The positive and negative losses are then computed as weighted sums over these masks:
\begin{equation}
\label{eq:pos_loss}
    \mathcal{L}_\text{pos} = - \frac{1}{Z} \sum_{i,j} \boldsymbol{M}^+_{ij} \, \boldsymbol{W}_{ij} \, \boldsymbol{C}_{\boldsymbol{F},ij},
\end{equation}
\begin{equation}
    \label{eq:neg_loss}
    \mathcal{L}_\text{neg} = \frac{1}{Z} \sum_{i,j} \boldsymbol{M}^-_{ij} \, \boldsymbol{W}_{ij} \, \mathrm{ReLU}(\boldsymbol{C}_{\boldsymbol{F},ij}),
\end{equation}
where $Z$ is a normalization factor (e.g., the total number of pixel pairs), $\boldsymbol{W} \in \mathbb{R}^{N_p \times N_p}$ is a pixel-weight matrix similar to \cite{cen2023saga}, and $\mathrm{ReLU}(\cdot)$ ensures that only positive contributions are considered for negative pairs. 
The final loss is then defined in \cref{eq:final_loss} with an extra regularization term to stabilize the feature learning. 
\begin{align}
\label{eq:final_loss}
\mathcal{L}&=\mathcal{L}_\text{pos}+\mathcal{L}_\text{neg}+\mathcal{L}_\text{reg}, \\
\mathcal{L}_{\text{reg}} &= \left( 1 - \frac{1}{HW} \sum_{i=1}^{HW} \lVert \boldsymbol{I}_{F}(u_i, v_i) \rVert_2 \right)^2
\end{align}

\noindent\textbf{Inference.}
To enable interactive applications with the learned 4D representation, we deploy DBSCAN algorithm \cite{ester1996density} to cluster the Gaussian features into several groups and generate the so-called Gaussian clusters. In contrast to the K-Means algorithm~\cite{lloyd1982least, macqueen1967some}, it does not require any prior knowledge of the number of clusters in a scene.
As can be seen in~\cref{fig:teaser}, clusters represent distinct objects, which verifies accurate alignment of our learned feature space with the object-level semantic field.

We further observe that the object boundaries can be effectively refined by applying a simple similarity threshold, which prunes Gaussians with features that are less similar to the corresponding cluster's mean. This simple technique enhances the method's performance in challenging segmentation scenarios, such as those involving occlusions and affected by sparse observations.

%% file: sec/4_experimental.tex
\section{Experimental Results}
\label{sec:experimental}
We first provide details of our proposed 4D segmentation benchmark in~\cref{subsec:segbench} and outline the evaluation protocol in~\cref{subsec:baselines} . 
We present segmentation results of our model, along with both quantitative and qualitative comparisons to other related works \cite{cen2023saga, ye2023gaussian, labe2024dgd} in~\cref{subsec:results}. Last but not least, we demonstrate the capabilities of our learned semantic representation on a number of downstream tasks, including scene editing and selection via our designed graphical user interface (\cref{subsec:editing}). Please refer to the supplementary material for implementation details.
\subsection{4D Segmentation Benchmark}\label{subsec:segbench}
We manually annotate dynamic sequences of existing novel-view synthesis datasets and present their segmentation-aware versions: \textit{NeRF-DS-Mask}, \textit{HyperNeRF-Mask}, \textit{Neu3D-Mask}, \textit{Immersive-Mask}, and \textit{Technicolor-Mask}. Utilizing the powerful SAM2 model \cite{ravi2024sam2}, which provides consistent object masks across video sequences, we define the objects of interest for each sequence to evaluate their segmentation performance in our model and manually refine them. Qualitative examples of the augmented sequences, together with a description of the raw datasets, can be found in the supplementary material.
\begin{table*}
  \centering
  \scalebox{0.8}{
  \begin{tabular}{c c c c c c c c c c c}
    \toprule
      & \multicolumn{2}{c}{NeRF-DS-Mask} & \multicolumn{2}{c}{HyperNeRF-Mask} & \multicolumn{2}{c}{Neu3D-Mask} & \multicolumn{2}{c}{Immersive-Mask} & \multicolumn{2}{c}{Technicolor-Mask}\\ 
      Method & mIoU$\uparrow$ & mAcc$\uparrow$ & mIoU$\uparrow$ & mAcc$\uparrow$ & mIoU$\uparrow$ & mAcc$\uparrow$ & mIoU$\uparrow$ & mAcc$\uparrow$ & mIoU$\uparrow$ & mAcc$\uparrow$\\
    \midrule
      GG \cite{ye2023gaussian} & 0.8351&	0.9729& 0.8341	&0.9771& 0.8864*	&0.9932*& 0.7673*&	0.9776*& 0.7979*&	0.9800*\\
      SAGA \cite{cen2023saga} & 0.8072&	0.9644& 0.7660&	0.9579& 0.6941	&0.9516 &0.7395&	0.9747 &0.7913	&0.9792\\
      DGD \cite{labe2024dgd} & 0.7125&	0.9551& 0.8297	&0.9777& 0.7721	&0.9851 &0.7981&	0.9850 &0.7791&	0.9728\\
      SA4D \cite{ji2024segment} & 0.6740 & 0.9360 & 0.7371 & 0.9616 & 0.8832* & 0.9391* & 0.7987* & 0.9835* & 0.8271* & 0.9734* \\
      CGC \cite{contrastive_gaussians_2024} & 0.8100	& 0.9718& 0.8157&0.9768& 0.8754&0.9927 &0.8856	&0.9925 &0.8806&0.9812\\
      OpenGaussian \cite{wu2025opengaussian} & 0.6939	&0.9564 & 0.6311	&0.9411& 0.8178 & 0.9899 &0.7120	&0.9793&0.8390&	0.9694\\
      \acronym{} (Ours) & \textbf{0.8768}&	\textbf{0.9831}& \textbf{0.8663}	&\textbf{0.9845} &\textbf{0.9022}	&\textbf{0.9945} &\textbf{0.9234}&	\textbf{0.9945}& \textbf{0.9308}	&\textbf{0.9917}\\
    \bottomrule
  \end{tabular}
  }
  \caption{Segmentation accuracy of 4D segmentation methods on our benchmark. \acronym{} outperforms the baselines on average mIoU and mAcc. The quantitative results for each sequence can be found in the supplementary. *: GG and SA4D need consistent object IDs from a video tracker for supervision~\cite{cheng2023tracking}. To evaluate on the multi-view camera dataset, we train models on a camera closest to the test view.}
  \label{tab:avg_quantitative}
\end{table*}
\begin{figure*}[t]
  \centering
  \begin{overpic}[width=1\textwidth]{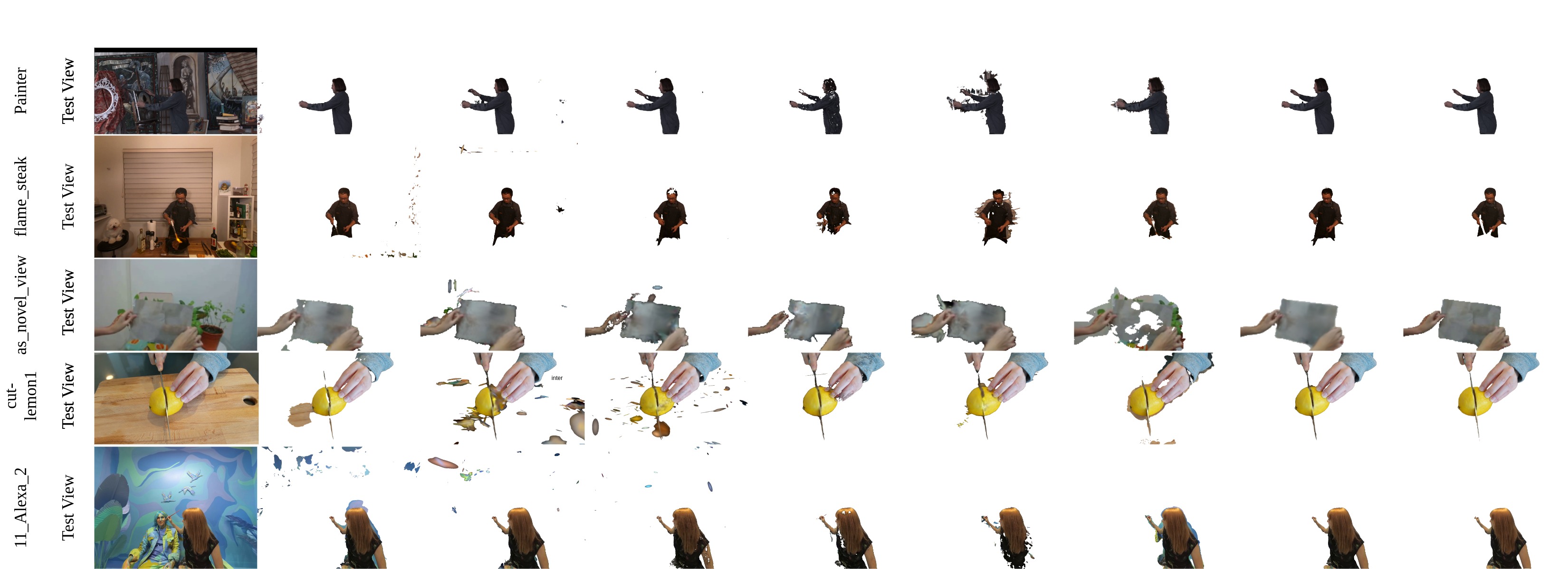}
    \put(8, 35){\scriptsize \colorbox{white}{\shortstack{GT Image}}}
    \put(18, 35){\scriptsize \colorbox{white}{\shortstack{Gaussian \\ Grouping}}}
    \put(29, 35){\scriptsize \colorbox{white}{\shortstack{SAGA}}}
    \put(40, 35){\scriptsize \colorbox{white}{\shortstack{CGC}}}
    \put(48, 35){\scriptsize \colorbox{white}{\shortstack{OpenGaussian}}}
    \put(61, 35){\scriptsize \colorbox{white}{\shortstack{DGD}}}
    \put(71, 35){\scriptsize \colorbox{white}{\shortstack{SA4D}}}
    \put(81, 35){\scriptsize \colorbox{white}{\shortstack{\acronym \\ (Ours)}}}
    \put(91, 35){\scriptsize \colorbox{white}{\shortstack{GT Object}}}
  \end{overpic}
  \caption{Segmentation qualitative results. While Gaussian Grouping \cite{ye2023gaussian} and SAGA \cite{cen2023saga} suffer from spurious Gaussians, our method demonstrates crisp and floater-free segmentation. Object boundaries in SA4D \cite{ji2024segment} and DGD \cite{labe2024dgd} segmentations are not tight and capture part of the background while rendering. 
  OpenGaussian \cite{wu2025opengaussian} is not able to fully capture the whole objects in its segmentation. 
  Our model consistently demonstrates superior segmentation quality and crisp object masks.
  } 
  \label{fig:qualitative}
  \vspace{-10pt}
\end{figure*}

\subsection{Evaluation Protocol and Baselines}\label{subsec:baselines}

We compare our model with SAGA \cite{cen2023saga}, Gaussian Grouping (GG) \cite{ye2023gaussian}, SA4D \cite{ji2024segment}, and DGD \cite{labe2024dgd}. Additionally, we add evaluation with the contrastive learning methods such as Contrastive Gaussian Clustering (CGC)~\cite{silva2024contrastive} and OpenGaussian~\cite{wu2025opengaussian}. Note that SAGA, Gaussian Grouping, CGC, and OpenGaussian are designed for static scenes and do not include a module to encode temporal information. To evaluate them on the 4D segmentation task, we integrate our 4D reconstruction backbone into their pipeline, extending their approach to effectively handle dynamic scenes. GG and SA4D require video object tracking labels for their supervision. To enable evaluation of the multi-view sequences from our benchmarks, we train the models on the camera closest to the test view (cam5 for Neu3D, cam2 for Immersive, and cam1 for Technicolor). 
After training, we manually perform click prompts in the rendered novel views of each scene trained by different approaches to select the objects of interest defined in the benchmark. Apart from the quantitative results, we also present qualitative results from various benchmarks, as certain insights cannot be fully captured through numerical evaluations alone. Following common practices, we use Mean Intersection over Union (mIoU) and Mean Pixel Accuracy (mAcc) to quantitatively evaluate the performance of the models.
\subsection{Segmentation Results}\label{subsec:results}
\begin{table}[ht]
  \centering
  \scalebox{0.80}{
  \begin{tabular}{c c c c || c}
    \toprule
      & Preproc & Training & Total (min) & Storage (MB)\\ 
    \midrule
      DGD \cite{labe2024dgd} & 0.08 & 103 & 103 & 285\\
      \acronym{} (Ours) & 16 & 16 & \textbf{32} & \textbf{60}\\
    \bottomrule
  \end{tabular}
  }
  \caption{Storage and time comparison between DGD \cite{labe2024dgd} and \acronym{} on NeRF-DS \cite{yan2023nerf}. \acronym{} achieves over 3$\times$ faster training time compared to DGD while only requiring 4.5$\times$ less memory in storage due to its compact Gaussian features.}
  \vspace{-2em}
  \label{tab:training_time_analysis}
\end{table}
As shown in \cref{tab:avg_quantitative}, \acronym{} achieves state-of-the-art performance across all benchmarks, surpassing other segmentation methods. Please refer to the supplementary material for details on each sequence of the benchmark. Further, we select some sequences to illustrate the quality of the segmentation of each model in \cref{fig:qualitative}. These comparisons reveal that \acronym{} qualitatively surpasses existing models and exhibits multi-view consistent segmentation performance. Notably, \acronym{} maintains strong segmentation consistency to test camera views in multi-view sequences such as \textit{flame\_steak}, \textit{Painter}, and \textit{11\_Alexa\_2}. This robust performance can be attributed to the compact Gaussian features learned through the proposed Gaussian feature training. In contrast, DGD segments objects by comparing cosine similarity scores between features retrieved via click prompts. However, determining a suitable threshold that generalizes across various scenes is neither straightforward nor intuitive.
\acronym{} employs clustering based on the Gaussian features to group the Gaussians into several segments. As this operation is done in 3D space, \acronym{} achieves cross-view consistency and doesn't need to associate objects from different views. This advantage is evident in \textit{cut-lemon1} example in~\cref{fig:qualitative}, where wrongly associated object IDs can cause segmentation failure for \cite{ye2023gaussian}.

%% file: sec/5_ablations.tex
\section{Ablation Studies}
\label{sec:ablations}
\subsection{Time and Storage Analysis}
\label{subsec:time_analysis}
\noindent We perform a comprehensive time and storage analysis of \acronym{} and DGD~\cite{labe2024dgd} using an RTX 3080 and an Intel i7-12700 on the NeRF-DS \cite{yan2023nerf} dataset. The results are shown in \cref{tab:training_time_analysis}, where ``Pre-Proc'' refers to the generation of DINOv2 384-dim features for DGD and SAM masks for \acronym{}. Rendering 384-dim features into 2D is computationally inefficient, significantly slowing down DGD's training process. \acronym{} uses 32-dim Gaussian features, which can be rendered much more efficiently. As a result, \acronym{} achieves over \textbf{3$\times$} faster total time than DGD while only requiring \textbf{4.5$\times$} less space in storage.
\subsection{Soft Sample Mining}
\label{subsec:ablate_contrastive_learning}
\begin{figure}[t]
  \centering
  \begin{overpic}[width=0.47\textwidth]{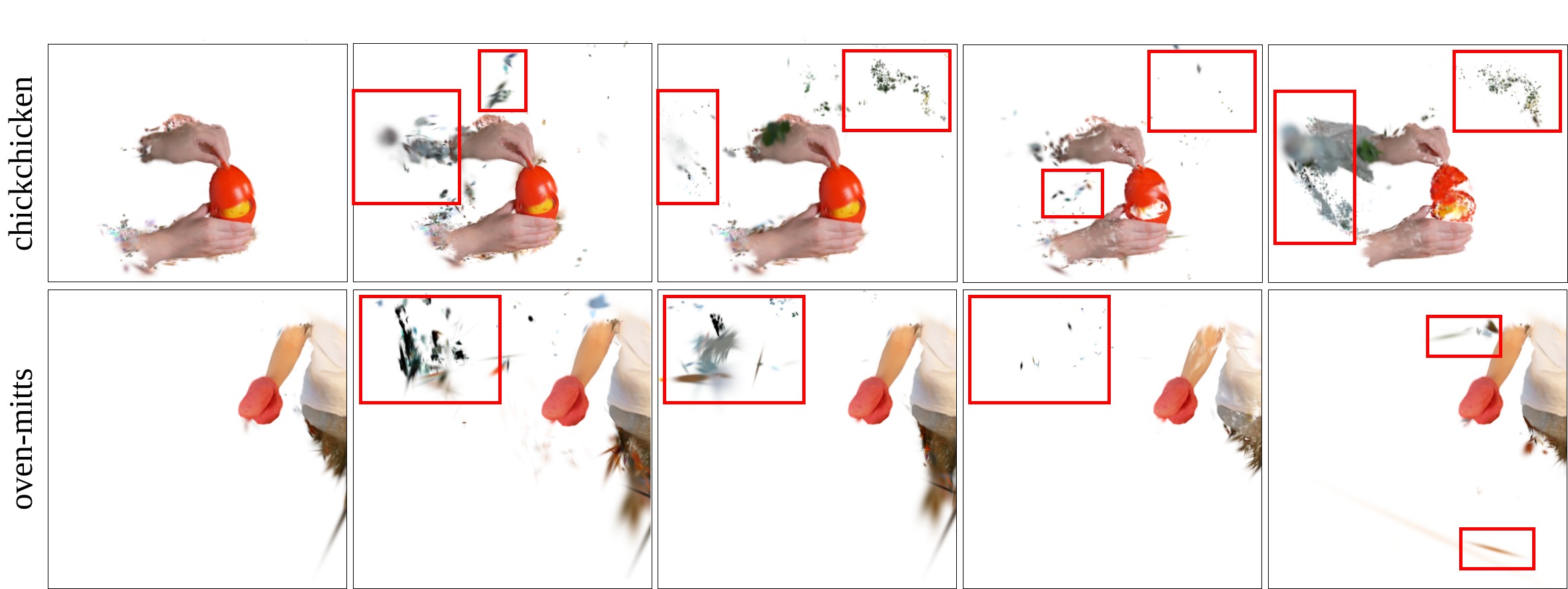}
  \put(6, 37){\fontsize{6}{7}\selectfont {\shortstack{\acronym{} (Ours)}}}
  \put(25, 37){\fontsize{6}{7}\selectfont {\shortstack{\acronym{} (hard)}}}
  \put(45, 37){\fontsize{6}{7}\selectfont {\shortstack{\acronym{} (all)}}}
  \put(68, 37){\fontsize{6}{7}\selectfont {\shortstack{CGC}}}
  \put(83, 37){\fontsize{6}{7}\selectfont {\shortstack{OpenGaussian}}}
  \end{overpic}
  \caption{Qualitative results of segmented objects in \textit{chickchicken} and \textit{oven-mitts} sequence in out-of-FoV viewpoint. \acronym{} with soft sample mining (Ours) exhibits superior performance over \acronym{} with hard / all mining, CGC \cite{silva2024contrastive}, and OpenGaussian \cite{wu2025opengaussian}.}
  \vspace{-15pt}
  \label{fig:out_of_fov_visualization}
\end{figure}

We visually demonstrate the effectiveness of our soft-mined pixel-pair sampling strategy in comparison to other contrastive objectives: sampling all positive–negative pairs (all) and sampling only pairs that individually satisfy the thresholds (hard). We include contrastive-based baselines such as CGC~\cite{silva2024contrastive} and OpenGaussian~\cite{wu2025opengaussian} for completeness. By investigating the out-of-FoV regions, i.e., 3D regions outside the observable space of train and test views, we observed floaters in the segmentation by our method without soft selection and other contrastive baselines, as shown in \cref{fig:out_of_fov_visualization}. Our mining approach effectively eliminates these floaters, leading to more stable and reliable segmentation. It is worth noting that no post-processing filtering is applied to ensure a fair comparison.
\subsection{Single-frame Mask Supervision}
Inspired by classical tracking methods~\cite{gladkova2022directtracker} and recent work linking 3DGS dynamics to pixel motion~\cite{gao2024gaussianflow}, we investigate the impact of incorporating temporal constraints on our single-frame mask supervision. Specifically, we compare our standard single-frame approach with a temporal extension, where masks are propagated across consecutive frames to encourage segmentation consistency over time. To implement this, we estimate optical flow between consecutive training timestamps $t_s$ and $t_{s+1}$ using differentiable rendering with accumulated 2D Gaussian displacements~\cite{gao2024gaussianflow}, and warp mask pixels from $t_s$ to $t_{s+1}$. Quantitative results are presented in~\cref{tab:flow_comp}. As observed, the multi-frame baseline does not provide significant improvements over the single-frame supervision, despite the doubled training cost, as an additional 2D flow and all associated parameters are rendered. This validates our design choice to rely on single-frame mask supervision, which achieves strong segmentation performance efficiently.
\begin{table}
  \centering
  \scalebox{0.65}{
  \centering
  \begin{tabular}{c c c c c c c}
    \toprule
      & \multicolumn{2}{c}{HyperNeRF-Mask} & \multicolumn{2}{c}{Immersive-Mask} & \multicolumn{2}{c}{Technicolor-Mask}\\ 
      Method & mIoU$\uparrow$ & mAcc$\uparrow$ & mIoU$\uparrow$ & mAcc$\uparrow$ & mIoU$\uparrow$ & mAcc$\uparrow$\\
    \midrule
      \acronym{} (multi-frame) & 0.8636	& 0.9833 & 0.8968	& 0.9933 & \textbf{0.9309}	& 0.9913\\
      \acronym{} (Ours) & \textbf{0.8663} &	\textbf{0.9845} & \textbf{0.9234} & \textbf{0.9945} & 0.9308 & \textbf{0.9917}\\
    \bottomrule
  \end{tabular}
  }
  \caption{Ablation study on single-frame mask supervision. The multi-frame baseline propagates masks between consecutive frames using Gaussian motion flow to impose temporal constraints. While this increases training time by a factor of \textbf{2}, it does not provide significant additional supervision, confirming the effectiveness of single-frame mask supervision.}
  \vspace{-5mm}
  \label{tab:flow_comp}
\end{table}

\begin{table}[ht]
  \centering
  \scalebox{0.63}{
  \begin{tabular}{c c c c c c c c}
    \toprule
      & & \multicolumn{2}{c}{NeRF-DS-Mask} & \multicolumn{2}{c}{HyperNeRF-Mask} & \multicolumn{2}{c}{Neu3D-Mask}\\ 
      Method & Filtering & mIoU$\uparrow$ & mAcc$\uparrow$ & mIoU$\uparrow$& mAcc$\uparrow$ & mIoU$\uparrow$ & mAcc$\uparrow$\\
    \midrule
    \multirow{2}{*}{SAGA~\cite{cen2023saga}} & \ding{56}& 0.8072&	0.9644& 0.7660&	0.9579& 0.6941	&0.9516 \\
    & \ding{51} & 0.8421&0.9778& 0.8223	& 0.9779& 0.8026&0.9853 \\
      \acronym{} & \ding{56} & 0.8760&	0.9829 &0.8368&	0.9790 &0.8738&	0.9923 \\
      \acronym{} (Ours) & \ding{51} &\textbf{0.8768}&	\textbf{0.9831}& \textbf{0.8663}	&\textbf{0.9845} &\textbf{0.9022}	&\textbf{0.9945} \\
    \bottomrule
  \end{tabular}
  }
  \caption{Effect of filtering spurious Gaussians for SAGA \cite{cen2023saga} and \acronym{}. Despite the improved baseline's performance, \acronym{} maintains the superior segmentation quality due to its spatially consistent feature space.}
  \label{tab:saga_w_filtering}
\end{table}
\begin{figure}[ht]
  \centering
  \scalebox{0.7}{
  \begin{overpic}[width=0.65\textwidth]{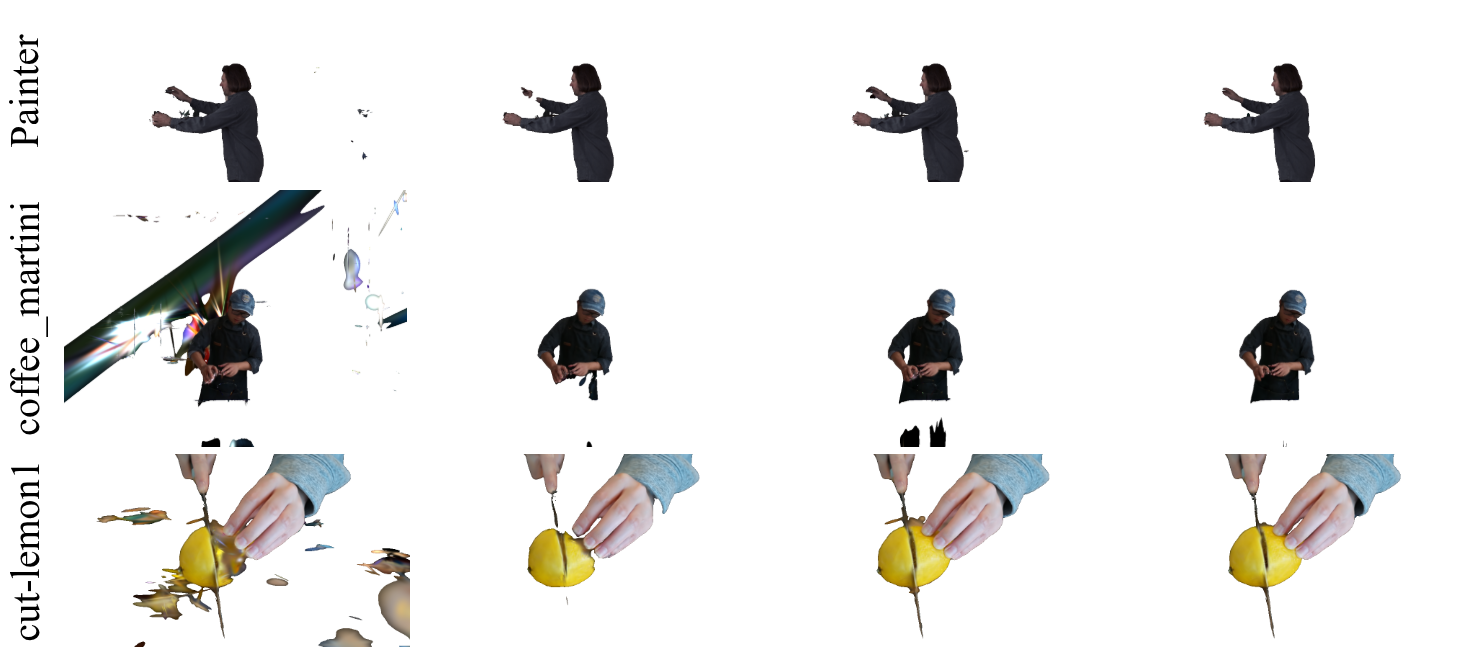}
  \put(9, 43){\fontsize{9}{7}\selectfont {\shortstack{SAGA \\w/o Filtering}}}
  \put(33, 43){\fontsize{9}{7}\selectfont {\shortstack{SAGA \\w/ Filtering}}}
  \put(57, 43){\fontsize{9}{7}\selectfont {\shortstack{\acronym{} \\w/o Filtering}}}
  \put(78, 43){\fontsize{9}{7}\selectfont {\shortstack{\acronym{} \\w/ Filtering (Ours)}}}
  \end{overpic}
  }
  \caption{Qualitative results for SAGA \cite{cen2023saga} with (w/) and without (w/o) our proposed filtering. Even though our filtering step can filter most of the artifacts from SAGA. Their resulting segmented objects are still not optimal compare to \acronym{}. }
  \vspace{-1.5em}
  \label{fig:saga_w_wo_filtering}
\end{figure}

\begin{figure*}[t]
  \centering
  \includegraphics[width=1\textwidth]{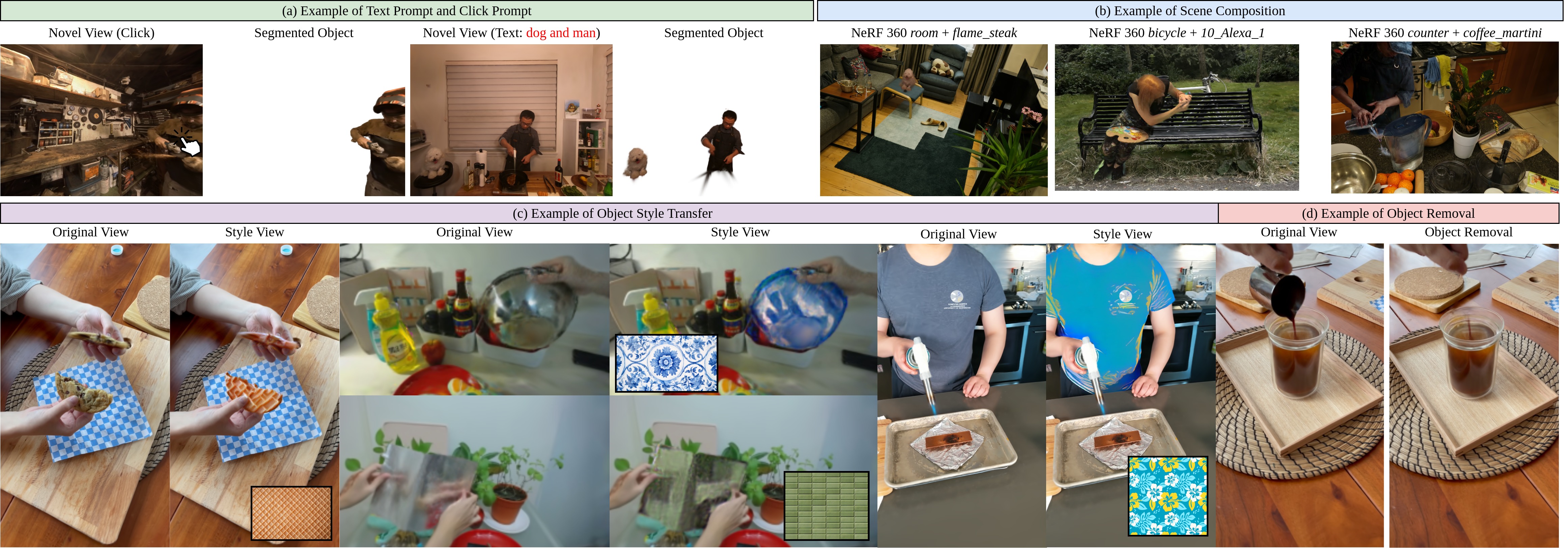}
  \caption{Versatile scene editing applications. (a) The object of interest can be selected by click prompts or text prompts. (b) Scene composition can be done by manipulating the selected Gaussians in another scene. (c) Style transfer of the segmented objects to obtain different textures. (d) Object removal for the selected object.}
  \vspace{-5pt}
  \label{fig:3d_editing}
\end{figure*}

\subsection{Post-processing Filtering}
We provide an ablation study of our method with respect to the filtering step, which removes Gaussians whose features are too dissimilar to the object's mean. As shown in \cref{tab:saga_w_filtering}, the proposed filtering step enhances the segmentation quality of our method. Particularly, artifacts on the object's boundaries are effectively removed, mitigating the inaccuracies in input masks or the effects of occlusions.

We further demonstrate the advantage of our learned feature field by showing that while the post-processing step enhances segmentation accuracy, it is not the sole factor driving our superior performance. We also note that this technique is applicable only to certain baselines such as SAGA~\cite{cen2023saga}. For example, methods like CGC~\cite{silva2024contrastive} and DGD~\cite{labe2024dgd} already incorporate a similarity threshold during inference. OpenGaussian~\cite{wu2025opengaussian} gets the assigned class ID for each Gaussian from their codebook. By contrast, SA4D~\cite{ji2024segment} and Gaussian Grouping~\cite{ye2023gaussian} learn an identity field, making our post-processing inapplicable to them. In \cref{tab:saga_w_filtering} and \cref{fig:saga_w_wo_filtering}, we quantify and illustrate the effectiveness of applying our proposed filtering step to SAGA~\cite{cen2023saga}. However, because the features learned under the baseline’s hard-contrastive objective remain suboptimal, the segmentation quality still exhibits more artifacts compared to \acronym{}.

%% file: sec/6_downstream_apps.tex
\section{Editing Applications}\label{subsec:editing}
In addition to accurate segmentation capabilities and multi-view consistent rendered masks, we demonstrate the effectiveness of our learned semantic feature field in a range of downstream tasks. Specifically, we focus on interactive editing applications and develop a graphical user interface (GUI) to grant full user control over the representation. Thanks to its explicit nature and our semantic abstraction to Gaussian clusters, we achieve real-time operation capability in handling many editing tasks. We provide snapshots of the GUI in the supplementary.

In addition to simple click prompts, we also support text-based prompts to make object selection even more intuitive. To enhance 3D content editing in dynamic scenes, we integrate style transfer capabilities for the selected objects. By segmenting objects directly in 3D, we facilitate scene composition and object removal through straightforward manipulation of the selected Gaussians. Due to the space limit, we shall provide more qualitative results in the supplementary. In the following, we describe the implementation details of editing tasks based on our semantically-aware feature field.

\noindent \textbf{Text Prompt.} Inspired by Gaussian Grouping \cite{ye2023gaussian}, we employ Grounded SAM~\cite{ren2024gsam} to generate a 2D mask based on the text prompt in the novel view. The 2D mask is reprojected into the 3D space with the rendered depth information. Finally, we associate each reprojected point with its nearest Gaussian in the 3D scene and retrieve its corresponding cluster. If the number of associated Gaussians of a given cluster is more than a threshold, the cluster would be selected. The example of text prompts on different sequences is shown in~\cref{fig:3d_editing} (a).

\noindent \textbf{Scene Composition.} 
We also perform qualitative results on scene composition. As the segmentation is performed in 3D, we can simply put the selected Gaussians from the dynamic scene into another static or dynamic scene. The scales and coordinate transformation need to be adapted. We show some examples of scene composition in~\cref{fig:3d_editing}.

\noindent \textbf{Object Style Transfer.}
We adopt the static Gaussian style transfer from StyleSplat \cite{jain2024stylesplat} for dynamic scenes. Given a rendered image $\boldsymbol{I}_r$ and a style image $\boldsymbol{I}_s$, we extract their feature maps ($\boldsymbol{F}_r$, $\boldsymbol{F}_s$ respectively) from VGG16~\cite{SimonyanZ14a} and optimize only the SHs of the Gaussians of the selected cluster with the nearest-neighbor feature matching (NNFM) loss \cite{zhang2022arf}. 
\cref{fig:3d_editing} (c) illustrates an example of style transfer on the segmented object.

\noindent \textbf{Object Removal.} 
We can also remove objects by simply deleting Gaussians belonging to the specific cluster as illustrated in \cref{fig:3d_editing} (d). It is worth noting that an additional inpainting step may be required, where the newly exposed parts have not been learned from multi-view observations.

%% file: sec/7_conclusion.tex
\section{Conclusion}
\label{sec:conclusion}
We introduced a novel framework for dynamic scene understanding, which enables multi-view consistent segmentation without any object tracking supervision. Our \acronym{} effectively combines dynamic 3D Gaussian Splatting 3DGS~\cite{yang2024deformable} and 2D SAM~\cite{kirillov2023sam} masks in a contrastive learning objective, which lifts semantic information into 3D space and learns expressive Gaussian features based on soft sample mining. This results in cross-view consistency when rendering segmented objects, enhancing the quality and coherence of object segmentation across different views. 
Evaluated on various novel-view datasets, \acronym{} shows superior performance both quantitatively and qualitatively. We further demonstrate the effectiveness of the learned feature field on downstream editing tasks such as point and text prompts, style transfer, object removal, and scene composition. Our approach sets a strong foundation for further research into dynamic scene understanding and scene editing, especially in complex multiview scenarios. We intend to release our code and benchmarks for future developments.

%% file: sec/X_suppl.tex
\clearpage
\setcounter{page}{1}
\maketitlesupplementary

In the supplementary, we provide several examples from our segmentation benchmark, along with dataset details (\cref{sec:segmentation_benchmarks}). Further implementation details are described in~\cref{sec:implementations} and \cref{sec:test_implementations}. 
A range of ablation studies on different components of our pipeline is given in~\cref{sec:ablate}. In~\cref{sec:supp_limitations} we address several limitations of our method and potential directions for future work. We provide additional visualizations of our learned feature space in comparison to other contrastive-based segmentation methods in \cref{sec:supp_feat_space_comparison} and \cref{sec:supp_feat_space}. In \cref{sec:supp_outfov} we show more examples on how our soft mining successfully mitigates floaters. Further, we present detailed quantitative results on all benchmarks in~\cref{sec:quant_res}. We finally demonstrate qualitative results of our method on segmentation (\cref{sec:further_qualitative}, \cref{sec:visualization_gaussian_clusters}) and editing (\cref{sec:editing}).
\section{Segmentation Benchmark}
\label{sec:segmentation_benchmarks}
\begin{figure*}[ht]
  \centering
  \includegraphics[width=1\textwidth]{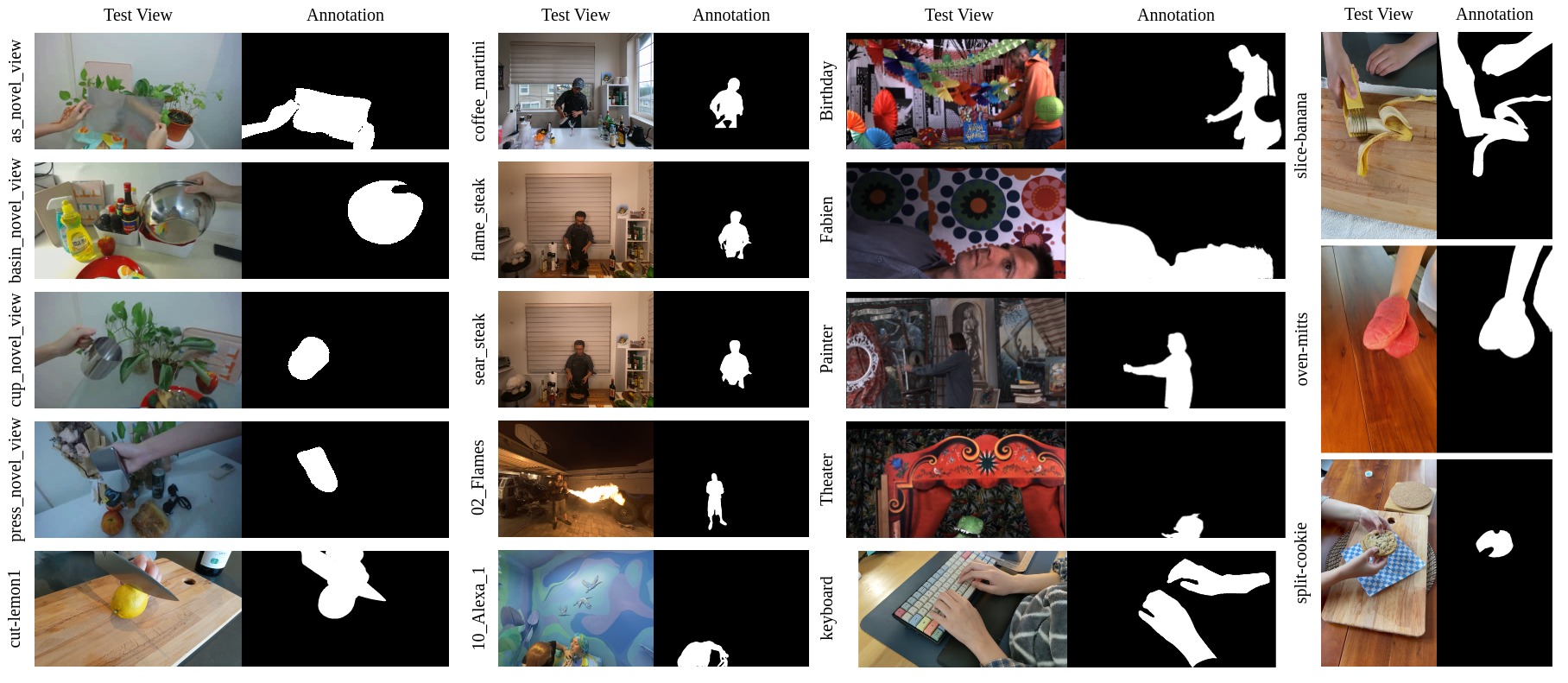}
  \caption{Image-object mask examples from our proposed extension to existing NVS datasets for dynamic 4D segmentation and downstream editing tasks. We ensure multi-view and temporal consistencies of the masks. }
  \label{fig:mask_benchmarks}
\end{figure*}

As discussed in Sec. 4.1 (main), we employ SAM2 \cite{ravi2024sam2} to manually select objects of interest to form our segmentation benchmarks for dynamic scene novel view synthesis. The examples are illustrated in \cref{fig:mask_benchmarks}. We describe the details of the mask storage for training and datasets adapted for our segmentation benchmark in the following.\\
\textbf{Mask Storage.} We employ SAM \cite{kirillov2023sam} to generate the anything-masks for Gaussian feature learning. However, the resulting masks are memory-inefficient because they are stored in a dense format ($M \times W \times H$) with a lot of zero-pixels. To optimize memory usage, we save the masks as a bit array. Given a permuted tensor version of an RGB image $I \in \mathbb{R}^{3 \times W \times H}$, SAM generates masks $\boldsymbol{\mathcal{M}} \in \mathbb{R}^{M \times W \times H}$. We first flatten $\boldsymbol{\mathcal{M}}$, then convert flattened ($M \times W \times H$)-dimensional array into a bit-array. The final output is stored in a dictionary with the required information to re-create the original mask tensor: a number of masks $M$, image dimensions $(W, H)$, and the compressed bit array, i.e., ${M, W, H, \text{bitarray}}$. This approach significantly reduces the storage used and is particularly useful for multi-view datasets, where numerous masks from different views are generated. For instance, a raw mask generated from an image in \textit{coffee\_martini} occupies around \textbf{80 MB}, but the dictionary representation reduces to about \textbf{1 MB}.\\
\noindent\textbf{NeRF-DS.} \cite{yan2023nerf} contains scenes with challenging transparent and shiny moving objects recorded by two monocular cameras. Five sequences are used for reconstruction and evaluation including \textit{as\_novel\_view}, \textit{basin\_novel\_view}, \textit{cup\_novel\_view}, \textit{press\_novel\_view}, and \textit{plate\_novel\_view}.\\
\noindent\textbf{HyperNeRF.} \cite{park2021hypernerf} is recorded using a single monocular camera 
. The camera moves while capturing the dynamic scenes involving different human-object interactions. In our experiments, we use ten sequences: \textit{americano}, \textit{split-cookie}, \textit{oven-mitts}, \textit{espresso}, \textit{chickchicken}, \textit{hand1-dense-v2}, \textit{torchocolate}, \textit{slice-banana}, \textit{keyboard}, and \textit{cut-lemon1}.\\
\noindent\textbf{Neu3D.} \cite{li2022neural}  contains videos captured by a multi-view rig with $18$ to $21$ cameras each at 30 FPS. The scenes involve a person performing cooking tasks in the kitchen. In total, five scenes are used in our experiments: \textit{coffee\_martini}, \textit{cook\_spinach}, \textit{cut\_roasted\_beef}, \textit{flame\_steak}, and \textit{sear\_steak}.\\
\noindent\textbf{Google Immersive.} \cite{broxton2020immersive} consists of videos captured by a 46-camera rig at 30 FPS. The cameras are fish-eye cameras and are distributed on the surface of a hemispherical, 92cm diameter dome. Four scenes are used in our experiments: \textit{01\_Welder}, \textit{02\_Flames}, \textit{10\_Alexa\_1}, and \textit{11\_Alexa\_2}.\\
\noindent\textbf{Technicolor Light Field.} \cite{sabater2017dataset} captures scenes using a $4 \times 4$ camera array at 30 FPS. We use the undistorted images provided by the authors, and four scenes are used in our experiments: \textit{Birthday}, \textit{Fabien}, \textit{Painter}, and \textit{Theater}.

\section{Training Implementation Details}
\label{sec:implementations}
We implement the whole pipeline in PyTorch \cite{paszke2019pytorch}. We reuse the codebase from Deformable-3DGS \cite{yang2024deformable} and extend the differential Gaussian rasterization pipeline to render the Gaussian features similar to \cite{ye2023gaussian}. In the following sections, we present the detailed training and evaluation setup for various datasets. 

\subsection{NeRF-DS and HyperNeRF}

For the NeRF-DS sequences, we use the left camera for training and the right camera
for testing. In the HyperNeRF sequences, training frames are sampled every 4 frames,
starting from the first frame, while testing frames are selected every 4 frames, starting
from the third frame. This follows the interp-sequence data-loading pipeline from the
original HyperNeRF paper \cite{park2021hypernerf}, ensuring that all testing frames are unseen during training. Note that we use $2\times$-downsized images for all experiments.

\begin{table*}
  \centering
  \scalebox{0.8}{
  \begin{tabular}{c c c c c c c c c c c}
    \toprule
      & \multicolumn{2}{c}{NeRF-DS-Mask} & \multicolumn{2}{c}{HyperNeRF-Mask} & \multicolumn{2}{c}{Neu3D-Mask} & \multicolumn{2}{c}{Immersive-Mask} & \multicolumn{2}{c}{Technicolor-Mask}\\ 
      Method & mIoU & mAcc & mIoU & mAcc & mIoU & mAcc & mIoU & mAcc & mIoU & mAcc\\
    \midrule
    \acronym{} (K=0) & 0.8228&	0.9706 &0.8360&	0.9787 &0.8661&	0.9920 &0.8725&	0.9909& \textbf{0.9259}	&\textbf{0.9908}\\
      Ours (K=16) & \textbf{0.8760}&	\textbf{0.9829} &\textbf{0.8368}&	\textbf{0.9790} &\textbf{0.8738}&	\textbf{0.9923} &0.9154	&0.9942& 0.9258&	0.9904\\
      \acronym{} (K=32) & 0.8754&0.9828& 0.8331 &	0.9780&0.8702&0.9922&\textbf{0.9217}&\textbf{0.9944}& 0.9255&	0.9904\\
    \bottomrule
  \end{tabular}
  }
  \caption{Comparison of average mIoU and mAcc per benchmark with K=0 (no smooth Gaussian features), K=16, and K=32. Our approach strikes a balance between performance and runtime requirements.}
  \label{tab:ablation_smooth_k}
\end{table*}
\noindent\textbf{Dynamic Geometry Reconstruction.} 
\label{sec:nerfds_dynamic_geo}
We adopt the default hyper-parameters from Deformable-3DGS \cite{yang2024deformable}.
Initially, the Gaussians are warmed up without the deformation MLP for the first 3k iterations to better capture the positions and shape stably as suggested in \cite{yang2024deformable}, after which the MLP optimization begins, and the 3D Gaussian and the deformation MLP are trained jointly. The densification of Gaussians for adaptive density control introduced in 3DGS \cite{kerbl3Dgaussians} is set to run until 15k iterations. The dynamic geometry reconstruction is trained for a total of 20k iterations. 

\noindent\textbf{Gaussian Feature Learning.} After the dynamic reconstruction is completed, we freeze model's weights and begin Gaussian feature learning for an additional 10k iterations. For each image, we randomly sample 25 masks ($M^{\prime}=25$) generated by SAM \cite{kirillov2023sam} and 5k pixels ($N_p=5000$), with K = 16 to get smooth Gaussian features. 
\begin{figure}[ht]
  \centering
  \includegraphics[width=0.4\textwidth]{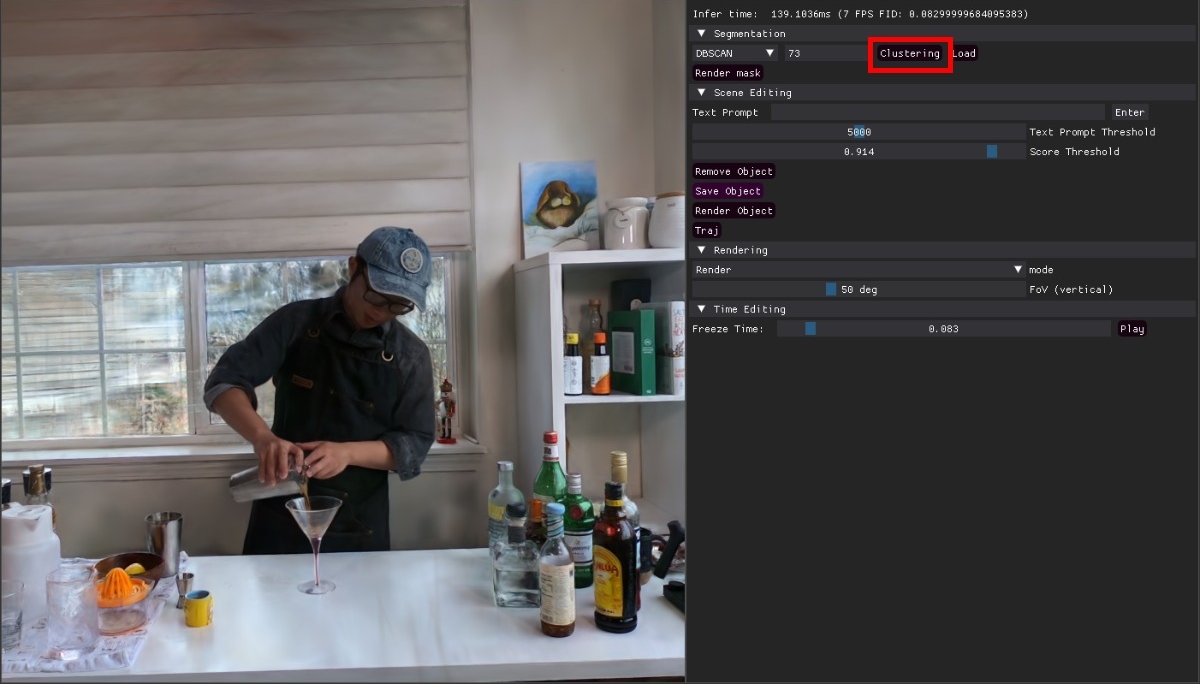}
  \caption{Example of our GUI operation. After opening the GUI, click ``Clustering'' to perform Gaussian feature clustering. }
  \label{fig:gui_overview}
  \vspace{-3mm}
\end{figure}
\begin{figure}[ht]
  \centering
  \includegraphics[width=0.4\textwidth]{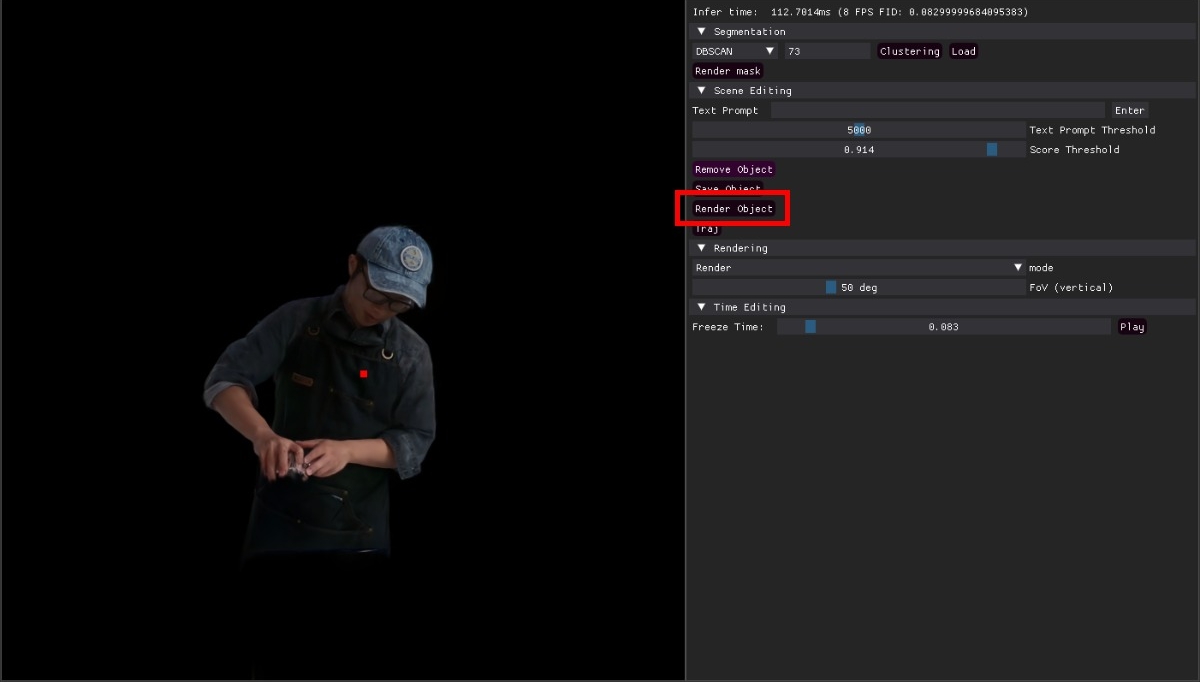}
  \caption{Example of our GUI operation. Hold the key ``A'' and click (\includegraphics[height=1.5ex]{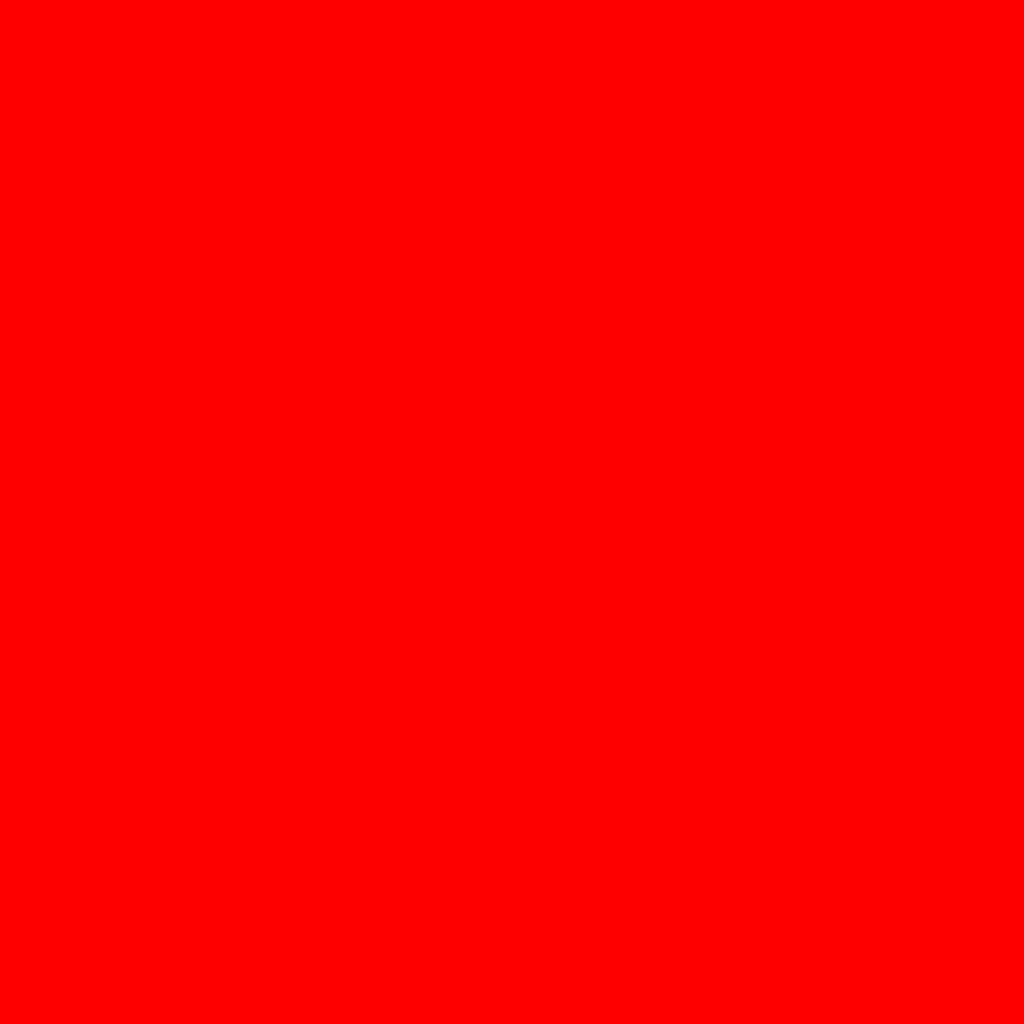}) on the objects of interest in the novel view to segment the target objects.}
  \label{fig:GUI_object}
  \vspace{-4mm}
\end{figure}
\subsection{Neu3D}
\label{sec:neu3D_implementation}
Each video contains 300 frames, and we begin by down-scaling the images by a factor of 2, resulting in a resolution of $1352 \times 1014$. The center camera (camera\_00) is used for testing, while the remaining cameras are for training. Initially, we run SfM~\cite{schoenberger2016sfm} across all the first frames, excluding camera\_00, to get the initial point cloud.

\noindent\textbf{Dynamic Geometry Reconstruction.}
We follow the same pipeline for dynamic geometry reconstruction as described in \cref{sec:nerfds_dynamic_geo}, except the densification of Gaussians is set to 8k iterations. 

\noindent\textbf{Gaussian Feature Learning.} 
After the dynamic reconstruction is completed, we freeze the model's weights and begin Gaussian feature learning for an additional 10k iterations.
For each image, we randomly sample 50 masks ($M^{\prime}=50$) generated by SAM \cite{kirillov2023sam} and 10k pixels ($N_p=10000$), with K = 16 to get the smooth Gaussian feature.

\subsection{Google Immersive}
Each video contains frames captured by fish-eye cameras with 300 frames.
We first undistort the images and align the principal points to the center, then downscale them to a resolution of $1280 \times 960$.

\begin{table*}
  \centering
  \scalebox{0.65}{
  \begin{tabular}{c c c c c c c c c c c c c c c c}
    \toprule
      & \multicolumn{3}{c}{NeRF-DS-Mask} & \multicolumn{3}{c}{HyperNeRF-Mask} & \multicolumn{3}{c}{Neu3D-Mask} & \multicolumn{3}{c}{Immersive-Mask} & \multicolumn{3}{c}{Technicolor-Mask}\\ 
    \midrule
      Avg. Num. Gaussians & \multicolumn{3}{c}{186455.4} & \multicolumn{3}{c}{565614.7} & \multicolumn{3}{c}{678566} & \multicolumn{3}{c}{985713} & 
      \multicolumn{3}{c}{457677.75}\\ 
    \midrule
      Method & mIoU & mAcc & time$\downarrow$ & mIoU & mAcc & time$\downarrow$ & mIoU & mAcc& time$\downarrow$ & mIoU & mAcc& time$\downarrow$ & mIoU & mAcc& time$\downarrow$\\
    \midrule
    *\acronym{} (2\%) &0.8760&0.9829&\textbf{0.4068}&0.8368&0.9790 &\textbf{2.3117} &0.8738&0.9923 &\textbf{2.2506}&0.9154	&0.9942&\textbf{9.6083} &0.9258&0.9904& \textbf{1.6118}	\\
      \acronym{} (10\%) &0.8707&0.9828 &1.5782&0.8384&0.9794 &11.8430&0.8792&0.9930 &6.1878&0.9160&0.9939&71.2429 &0.9283	&\textbf{0.9908}&6.8211\\
      \acronym{} (25\%) &\textbf{0.8762}&\textbf{0.9832}&3.9425& \textbf{0.8461}&\textbf{0.9809} &43.2682&\textbf{0.8886}&\textbf{0.9935}&18.8976&\textbf{0.9180}&\textbf{0.9943}&221.7642&\textbf{0.9297}&0.9907&19.0767\\
    \bottomrule
  \end{tabular}
  }
  \caption{Comparison of average mIoU, mAcc, and clustering time per benchmark with different sub-sampling of Gaussians 2\%, 10\%, and 25\%. *Our selection for our pipeline.}
  \label{tab:ablation_time}
\end{table*}

\begin{figure}[ht]
  \centering
  \includegraphics[width=0.45\textwidth]{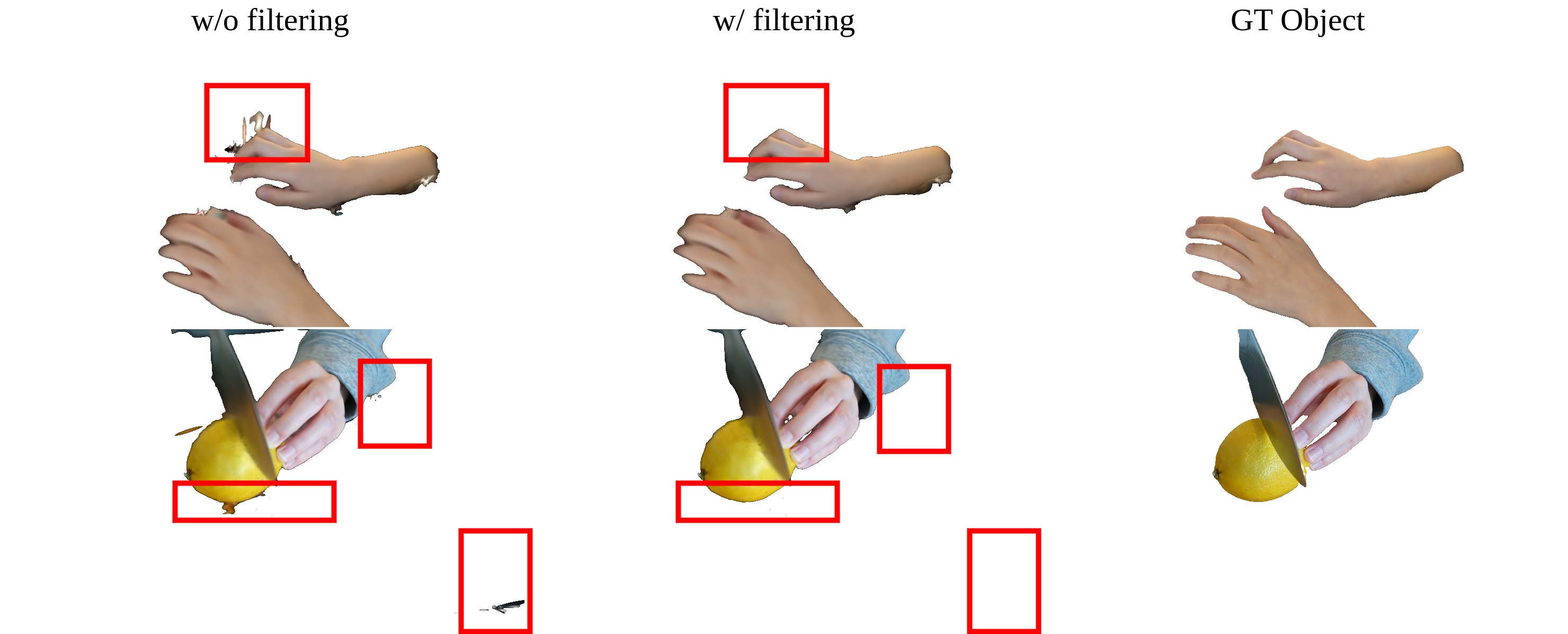}
  \caption{Visualization of ground-truth object of interest, segmented object without (w/o) filtering, and segmented object with (w/) filtering. }
  \label{fig:filtering}
\end{figure}

Similar to Neu3D, the center camera (camera\_0001) is used for testing, while the other cameras are used for training. We generate the initial point cloud using SfM across all the first frames, excluding the test camera, and apply the same hyper-parameters as in Neu3D in \cref{sec:neu3D_implementation}.

\subsection{Technicolor Light Field}
We use the undistorted images provided by the authors, aligning the principal points to the center and resizing the images to a resolution of $1024 \times 544$. 
Similar to Neu3D and Google Immersive dataset, the center camera (camera\_0000) is used for testing, while the other cameras are used for training. We generate the initial point cloud using SfM across all the first frames excluding the test camera and apply the same hyper-parameters as in Neu3D in \cref{sec:neu3D_implementation}.
\section{Testing Implementation Details}
\label{sec:test_implementations}
After the dynamic geometry reconstruction and Gaussian features are properly trained, we proceed to our GUI to select the object of interest and evaluate the model's performance. 

Given a clicked point $[u, v]^T$ on the novel view, we first use the known camera intrinsics to compute the inverse intrinsic matrix $\boldsymbol{K}^{-1}$. With the depth information $d$ rendered by the dynamic reconstruction of the Gaussians $\mathcal{G}_t$ at time $t$, we can reproject the 2D clicked point to 3D coordinates in the camera frame. Finally,
using the known camera extrinsics (rotation matrix $\boldsymbol{R}$ and translation vector $\boldsymbol{T}$) for the world-to-camera transformation, we convert the points back into the world coordinate frame. We can then use this point coordinate to query its nearest neighbor Gaussian using KNN. The cluster ID corresponding to its nearest neighbor Gaussian is subsequently retrieved.

We provide interactive illustrations in our GUI to segment target objects as follows:
\begin{enumerate}
    \item Open the GUI and run DBSCAN (Click button "Clustering") to group Gaussians into different clusters based on the learned Gaussian features as illustrated in \cref{fig:gui_overview}.
    \item Click the objects of interest in the novel view to segment the target objects as shown in \cref{fig:GUI_object}. 
    \item Optional: Adjust the threshold to refine the object's boundaries ($T_\text{Neu3D}=0.98$, $T_\text{NeRF-DS}=0.4$, $T_\text{HyperNeRF}=0.9$, $T_\text{Technicolor}=0.9$, $T_\text{Immersive}=0.6$).
    \item Click the button "Render Object" to render the selected object to test views for evaluating the performance. 
\end{enumerate}

\section{Ablation Studies}
\label{sec:ablate}

\subsection{Smooth Gaussian Features}
\label{sec:smooth_gaussian_features}
As mentioned in Sec. 3.2, we utilize smooth versions of Gaussian features, which can be defined as follows: 
\begin{equation}
    \boldsymbol{f}^{s}_{i} = \frac{1}{K} \sum_{j \in \text{KNN}(i)}\boldsymbol{f}_{j}.
    \label{eq:smooth_gaussian_feats}
\end{equation}
We provide a detailed analysis of the selection of the number of nearest neighbors for the KNN algorithm in \cref{tab:ablation_smooth_k}. Overall, when smooth Gaussian features are enabled ($K = 16$ and $K = 32$), the performance is better than when smooth Gaussian features are disabled ($K = 0$). Since there is no significant improvement between $K = 16$ and $K = 32$, we eventually chose $K = 16$ for our pipeline to avoid additional overheads with the KNN sampling.
\subsection{Gaussian Feature Clustering} 
\label{sec:gaussian_feature_clustering}
We also employ a subsampling strategy to enable real-time Gaussian feature clustering in scenes that typically contain more than 300k Gaussians. We perform time analysis with different sub-sampling numbers of Gaussians (2\%, 10\%, and 25\%) on RTX3080 and Intel i7-12700. For each scene, we compute the time for clustering using Python \textit{timeit} package and take the average clustering time for five runs per scene. We report the average clustering time across all scenes in each benchmark and the corresponding mIoU and mAcc for our object of interest in \cref{tab:ablation_time}. Overall, with different sub-sampling of Gaussians (2\%, 10\%, and 25\%), the mIoU  and mAcc stay almost unchanged. However, as increasing the number of Gaussians results in a longer clustering time, we believe that 2\% is the optimal sub-sampling number for clustering. 

\subsection{Sensitivity of Post-processing Filtering Threshold}
\begin{figure}[t]
  \centering
  \includegraphics[width=0.45\textwidth]{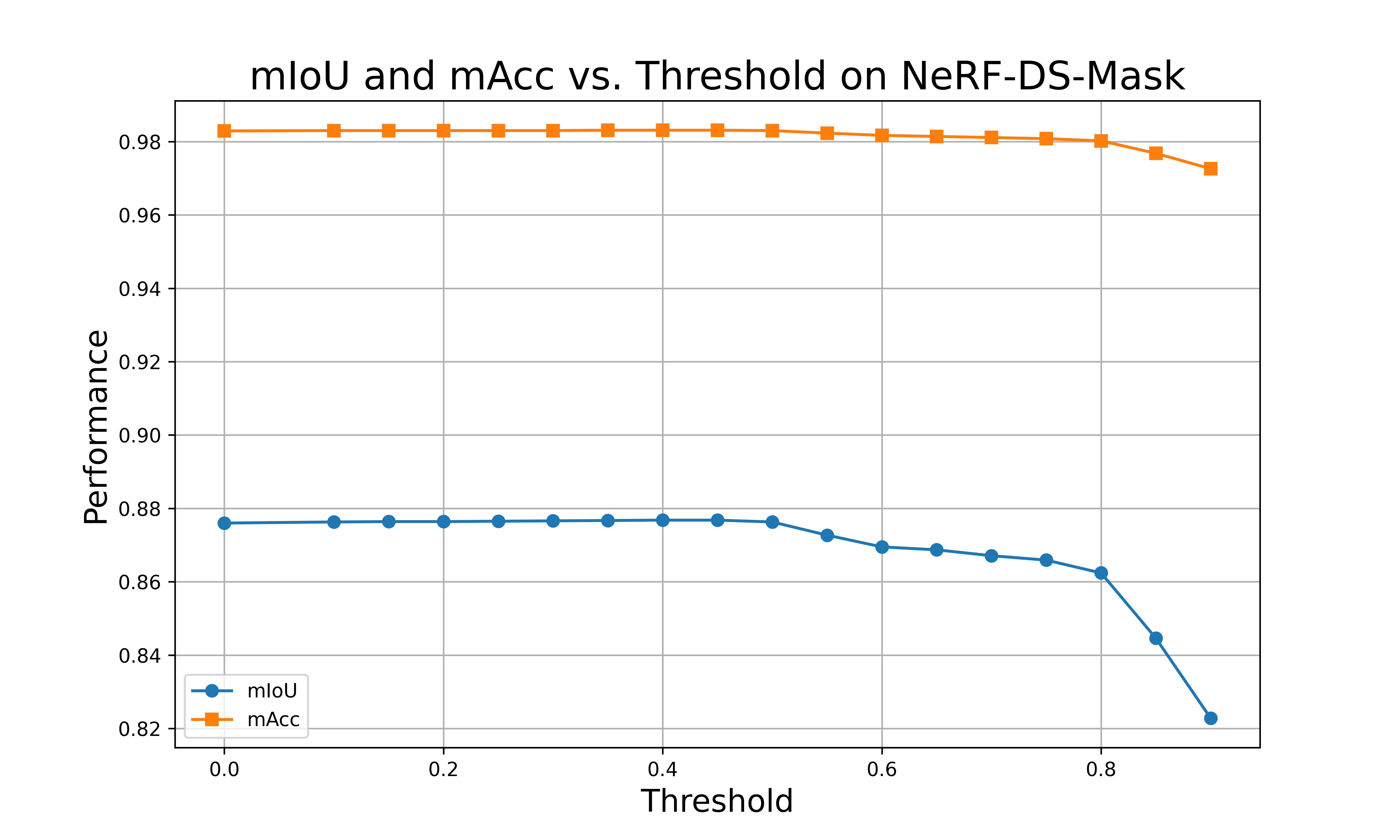}
  \caption{The sensitivity of \acronym{} with respect to the threshold selection for our post-processing filtering on NeRF-DS-Mask benchmark. Our model maintains accuracy across a wide range of threshold values, demonstrating the robustness and reliability of the learned feature field.}
  \vspace{-1em}
  \label{fig:sensitivity_th}
\end{figure}

We also evaluated the sensitivity of the post-processing filtering threshold. As illustrated in \cref{fig:sensitivity_th}, there exists a wide margin of effective values. This demonstrates that selecting a threshold is not difficult, as one can easily choose a valid setting by observing the performance tendency. 

\subsection{DBSCAN vs. K-Means}
\label{sec:dbscan_vs_kmeans}
There are various clustering techniques and K-means \cite{lloyd1982least, macqueen1967some} is one of the most widely used. K-means partitions data into $k$ clusters, where each feature belongs to the cluster with the nearest centroid (mean of the cluster). The value $k$ must be predefined, and the algorithm iteratively adjusts the centroids to minimize the sum of squared distances between features and their corresponding cluster centers. Therefore, the performance of the algorithm heavily depends on the optimal value of $k$.

\begin{figure*}[t]
  \centering
  \includegraphics[width=0.8\textwidth]{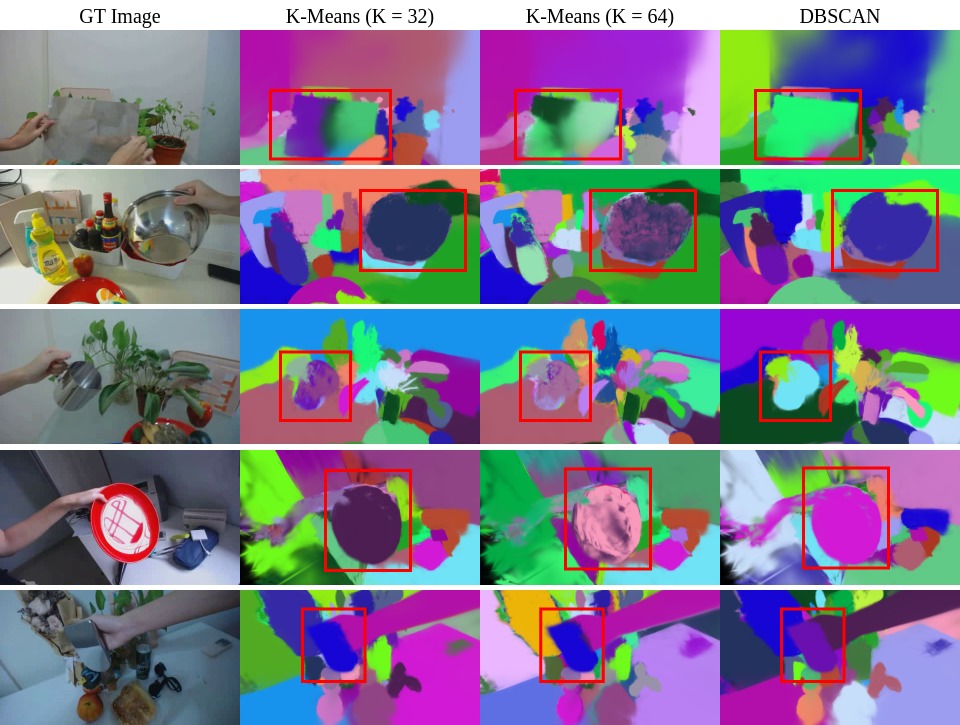}
  \caption{Comparison of rendered segmentation for K-Means (K=32), K-Means (K=64), and DBSCAN. K-Means-based segmentation results exhibit noise in the specular regions. Moreover, they require a pre-defined number of clusters, which is hard to provide for casual realistic videos. DBSCAN alleviates all of the above issues. \vspace{-2mm}}
  \label{fig:kmeans_dbscan}
\end{figure*}

To investigate this, we conducted an additional ablation study comparing the rendered segmentation produced by K-means with different values of $k$ and DBSCAN. As shown in \cref{fig:kmeans_dbscan}, the choice of $k$ significantly impacts the result of rendered segmentation. Since different sequences contain varying numbers of objects, we believe that employing a method with an adaptive number of clusters, such as DBSCAN, validates our design choices with additional flexibility and no loss on the segmentation accuracy.
\begin{table}
  \centering
  \scalebox{0.73}{
  \begin{tabular}{c c c c c c c}
    \toprule
      & \multicolumn{2}{c}{NeRF-DS-Mask} & \multicolumn{2}{c}{HyperNeRF-Mask} & \multicolumn{2}{c}{Neu3D-Mask}\\ 
      Sampling rule & mIoU & mAcc & mIoU & mAcc & mIoU & mAcc\\
    \midrule
      By Area & 0.8403 & 0.9743 & 0.8171 & 0.9632 & 0.8679 & 0.9920 \\
      By Score & 0.8028 & 0.9643 & 0.7998 & 0.9722 & 0.7673 & 0.9831 \\
      Grid-based & 0.8721 & 0.9822 & 0.8484 & 0.9809 & 0.8708 & 0.9922 \\
      Random (Ours) & \textbf{0.8768}&	\textbf{0.9831}& \textbf{0.8663}	&\textbf{0.9845} &\textbf{0.9022}	&\textbf{0.9945}\\
    \bottomrule
  \end{tabular}
  }
  \caption{Ablation study on mask sampling technique.}
  \label{tab:ablate_mask}
  \vspace{-10pt}
\end{table}
\subsection{Gaussian Feature Dimension}
\label{subsec:gaussian_feat_dim}
\begin{figure}[t]
  \centering
  \includegraphics[width=0.45\textwidth]{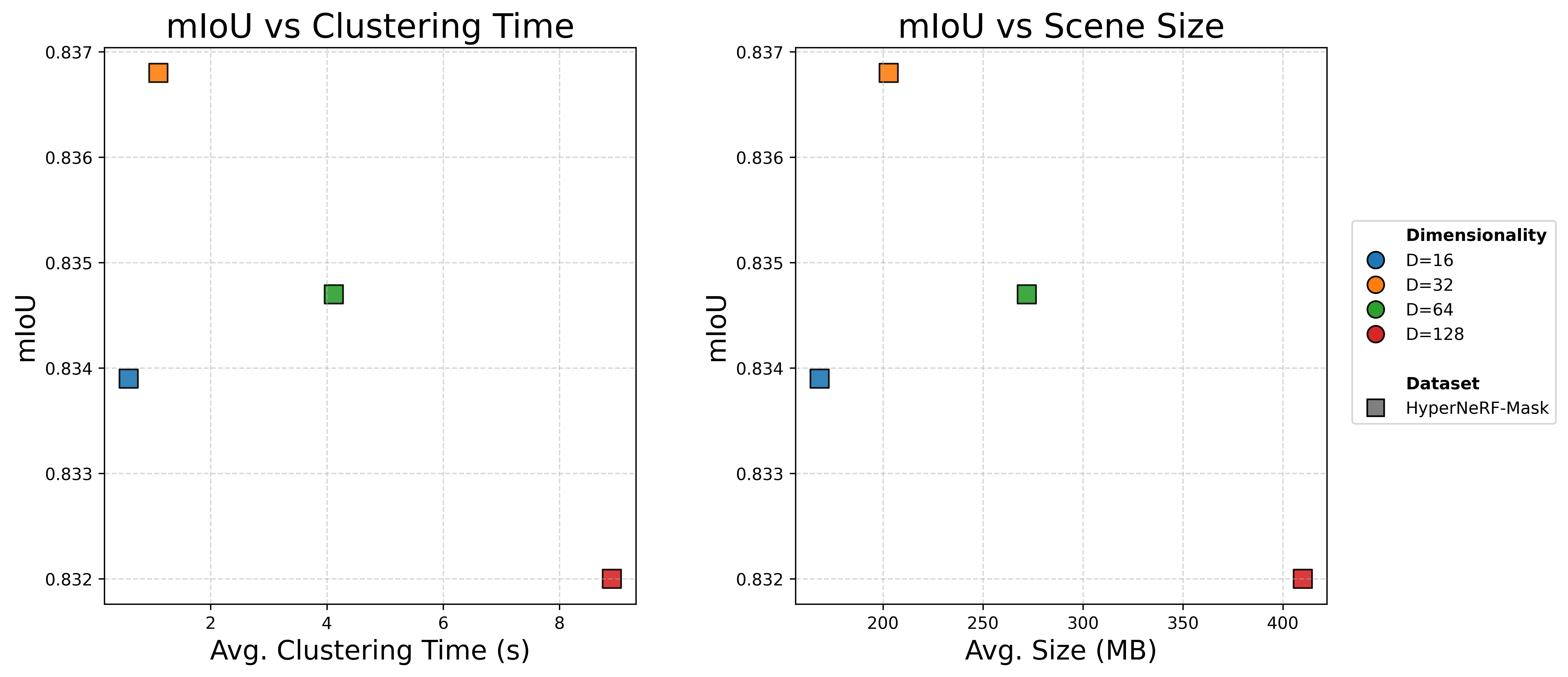}
  \caption{Scatter plot for the analysis of different Gaussian feature dimension $D\in\{16, 32, 64, 128\}$ on segmentation performance and runtime costs on HyperNeRF-Mask. \textbf{Left:} Comparison of mIoU against Average Clustering Time (seconds). \textbf{Right:} Comparison of mIoU against Average Scene Size (MB). The chosen configuration of $D=32$ strikes a balance between model accuracy and computational requirements.}
  \vspace{-1em}
  \label{fig:feature_dim_plot}
\end{figure}

We also analyze the impact of Gaussian feature dimensionality $D\in\{16, 32, 64, 128\}$ on segmentation performance. As shown in \cref{tab:feature_dim_analysis}, increasing $D$ does not lead to a clear improvement in mIoU or mAcc. 
On the other hand, the runtime costs such as storage requirements and clustering time increase significantly with larger $D$ as shown in \cref{tab:feature_dim_analysis_size_and_time}. \cref{fig:feature_dim_plot} visualizes these trade-offs by plotting the no-filtered mIoU from HyperNeRF-Mask against runtime costs. Ideally, the optimal configuration should be located at the top-left corner (high mIoU, low cost). The plot confirms that our chosen dimensionality ($D=32$) strikes the best balance between segmentation performance and computational efficiency.
\begin{table}[ht]
  \centering
  \scalebox{0.65}{
  \begin{tabular}{c l c c c c c c}
    \toprule
      & & \multicolumn{2}{c}{NeRF-DS-Mask} & \multicolumn{2}{c}{HyperNeRF-Mask} & \multicolumn{2}{c}{Neu3D-Mask}\\ 
      D & Filtering & mIoU$\uparrow$ & mAcc$\uparrow$ & mIoU$\uparrow$& mAcc$\uparrow$ & mIoU$\uparrow$ & mAcc$\uparrow$\\
    \midrule
    \multirow{2}{*}{D=16} & \ding{56}& 0.8730&0.9822& 0.8339&0.9778& 0.8702&0.9921\\
    & \ding{51} & 0.8733&0.9822& 0.8563&0.9828& 0.9036&0.9945 \\

    \multirow{2}{*}{D=32} & \ding{56}& 0.8760&0.9829& 0.8368&0.9790& 0.8738&0.9923\\
    & \ding{51} (Ours) & \textbf{0.8768}&\textbf{0.9831}& \textbf{0.8663}&\textbf{0.9845}& 0.9022&0.9945 \\

    \multirow{2}{*}{D=64} & \ding{56}& 0.8744&0.9825& 0.8347&0.9780& 0.8949	&0.9939\\
    & \ding{51} & 0.8756&0.9830& 0.8573&0.9832&0.9068&\textbf{0.9948}\\

    \multirow{2}{*}{D=128} & \ding{56}& 0.8752&0.9826& 0.8320&0.9775& 0.8936&0.9938\\
    & \ding{51} & 0.8761&0.9830& 0.8573	&0.9830&\textbf{0.9070}	&\textbf{0.9948}\\
      
    \bottomrule
  \end{tabular}
  }
  \caption{Performance analysis of different dimensionality of Gaussian features ($D\in\{16, 32, 64, 128\}$). }
  \label{tab:feature_dim_analysis}
\end{table}
\begin{table}[ht]
  \centering
  \scalebox{0.7}{
  \begin{tabular}{l c c c c c c c}
    \toprule
      & \multicolumn{2}{c}{NeRF-DS-Mask} & \multicolumn{2}{c}{HyperNeRF-Mask} & \multicolumn{2}{c}{Neu3D-Mask}\\ 
      D  & Size$\downarrow$ & Time$\downarrow$ & Size$\downarrow$ & Time$\downarrow$ & Size$\downarrow$ & Time$\downarrow$\\
    \midrule
    D=16 & \textbf{55}&\textbf{0.17}&\textbf{168}&\textbf{0.59}& \textbf{198}&\textbf{0.69}\\
    D=32 (Ours) & 67&0.30&203&1.10& 239&1.56\\
    D=64 & 90&0.43&272&4.12& 320&3.46\\
    D=128& 135&0.82&410&8.90& 483&7.06\\
    \bottomrule
  \end{tabular}
  }
  \caption{Clustering time (second) and storage size (MB) analysis of different dimensionality of Gaussian features ($D\in\{16, 32, 64, 128\}$).  \vspace{-1em}}
  \label{tab:feature_dim_analysis_size_and_time}
\end{table}
\subsection{Mask Sampling}
\label{subsec:mask_sampling}
We further study the influence of SAM~\cite{kirillov2023sam} masks sampling during the feature learning stage. We test three different sampling heuristics: spatially uniform, based on the mask area (number of pixels), and based on the mask score. For the latter, we consider the stability scores from SAM~\cite{kirillov2023sam} model, which reflects the reliability of the object against different object's boundary changes. We have noticed many partial segments receiving high scores, especially for objects with reflective surfaces from the NeRF-DS dataset. This negatively affects the segmentation accuracy for the model that prioritizes the masks with the highest stability scores. Spatially uniform masks are retrieved by dividing the image space into 50 cells and sampling a mask per cell. If several candidates correspond to one cell, masks are selected based on their area. This method generally works well. However, true uniform sampling highly depends on mask shape, size, and arrangement, which is hard to achieve with simple heuristics.
\section{Limitations and Future Work}
\label{sec:supp_limitations}
\begin{figure}[t]
  \centering
  \begin{overpic}[width=0.4\textwidth]{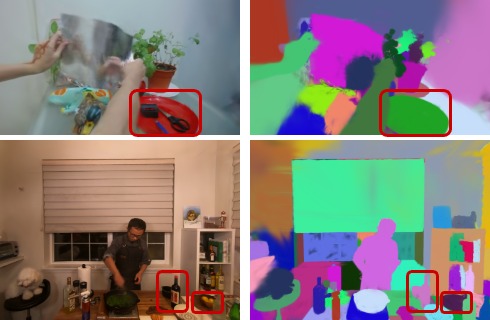}
  \put(-7, 37){\rotatebox{90}{as\_novel\_view}}
  \put(-7, 5){\rotatebox{90}{cook\_spinach}}
  \put(10, 67){Rendered View}
  \put(51, 67){Rendered Segmentation}
  \end{overpic}
  \caption{Examples of failure cases of \acronym{} on small and overlapping objects. }
  \label{fig:failure_cases_object_overlap}
\end{figure}
We observe that \acronym{} faces difficulties in separating small objects that significantly overlap with larger ones, as illustrated in \cref{fig:failure_cases_object_overlap}. Intuitively, the scale-aware training proposed by SAGA \cite{cen2023saga} should improve the segmentation of these smaller objects; however, our experiments indicate that their pipeline yields noisy segmentation results. A potential alternative is to refine the construction of the mask-based correspondence matrix $C$. While the current implementation encodes it with binary values (0 or 1), employing soft or probabilistic values could offer greater robustness.

As described in the main paper, our method follows a two-stage training scheme. Thus, exploring the benefits of learning dynamic reconstruction and segmentation jointly is an exciting future direction of \acronym{}. 
Moreover, due to the rapid advances in the field, novel dynamic 3DGS reconstruction backbones can be combined with our soft-mining feature learning to achieve better segmentation performance.
\section{Additional Evaluations of Soft-Mining Contrastive Objective}
\label{sec:supp_feat_space_comparison}
\noindent\textbf{Soft vs. Hard vs. All Mining. }To validate the effectiveness of our soft-mined pixel-pair sampling strategy, we compare its average intra-cluster and inter-cluster distances against two baselines: sampling all positive–negative pairs (all) and sampling only pairs that individually satisfy the thresholds (hard), as shown in \cref{tab:inter_intra}.
Our results show that soft sample mining reduces intra-cluster distances while increasing inter-cluster distances, thereby improving feature separability.
\begin{table}
  \centering
  \scalebox{0.65}{
  \begin{tabular}{c c c c c c c}
    \toprule
      & \multicolumn{2}{c}{NeRF-DS-Mask} & \multicolumn{2}{c}{HyperNeRF-Mask} & \multicolumn{2}{c}{Neu3D-Mask} \\ 
      Method & intra ($\downarrow$) & inter ($\uparrow$) & intra ($\downarrow$) & inter ($\uparrow$) & intra ($\downarrow$) & inter ($\uparrow$) \\
    \midrule
      \acronym{} (all) & 0.2588&	1.3756 &0.3251&	1.3382 &0.2200&	1.3894\\
      \acronym{} (hard) & 0.3517&	1.1027 & 0.3488 &	1.1078 & 0.2942 & 1.2600	\\
      \acronym{} (Ours) & \textbf{0.2538}&	\textbf{1.3761} &\textbf{0.2437}&	\textbf{1.3406} &\textbf{0.1752}	&\textbf{1.3925} \\
    \bottomrule
  \end{tabular}
  }
  \caption{Ablation study on our contrastive objective. We compare the mean intra- and inter-cluster distance with our soft mining strategy (Ours), sampling all positive–negative pairs (all) and sampling only pairs that individually satisfy the thresholds (hard). 
  }
  \vspace{-3pt}
  \label{tab:inter_intra}
\end{table}

\begin{figure}[t]
  \centering
  \begin{overpic}[width=0.45\textwidth]{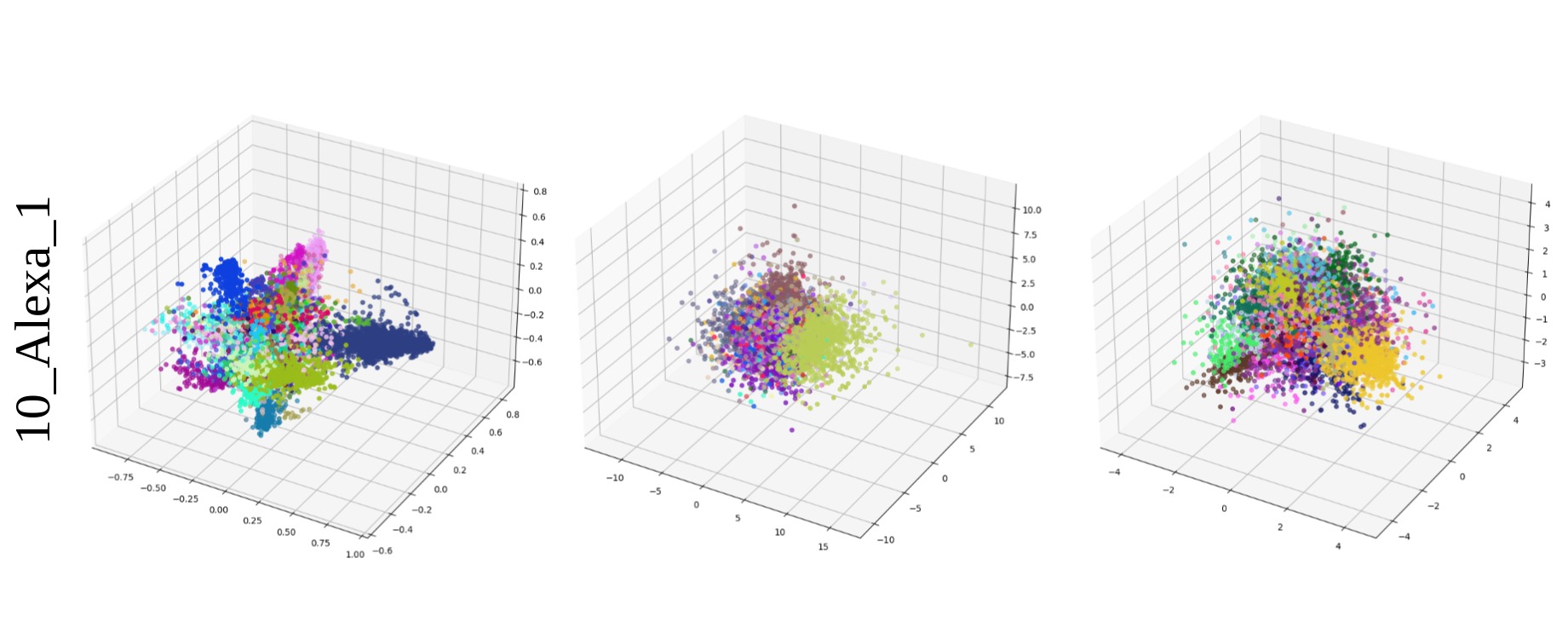}
      \put(13, 36){\scriptsize {\shortstack{\acronym}}}
      \put(47, 36){\scriptsize {\shortstack{CGC}}}
      \put(75, 36){\scriptsize {\shortstack{OpenGaussian}}}
  \end{overpic}
  \caption{Feature space visualization with PCA of CGC \cite{silva2024contrastive}, OpenGaussian \cite{wu2025opengaussian}, and \acronym{} on \textit{10\_Alexa\_1} sequence. \acronym{} exhibits clearer separation between different clusters and a directional structure, which allows to preserve more meaningful variations within a cluster.}
  \vspace{-3pt}
  \label{fig:feature_space_comparison}
\end{figure}

\begin{figure*}[t]
  \centering
  \begin{overpic}[width=0.8\textwidth]{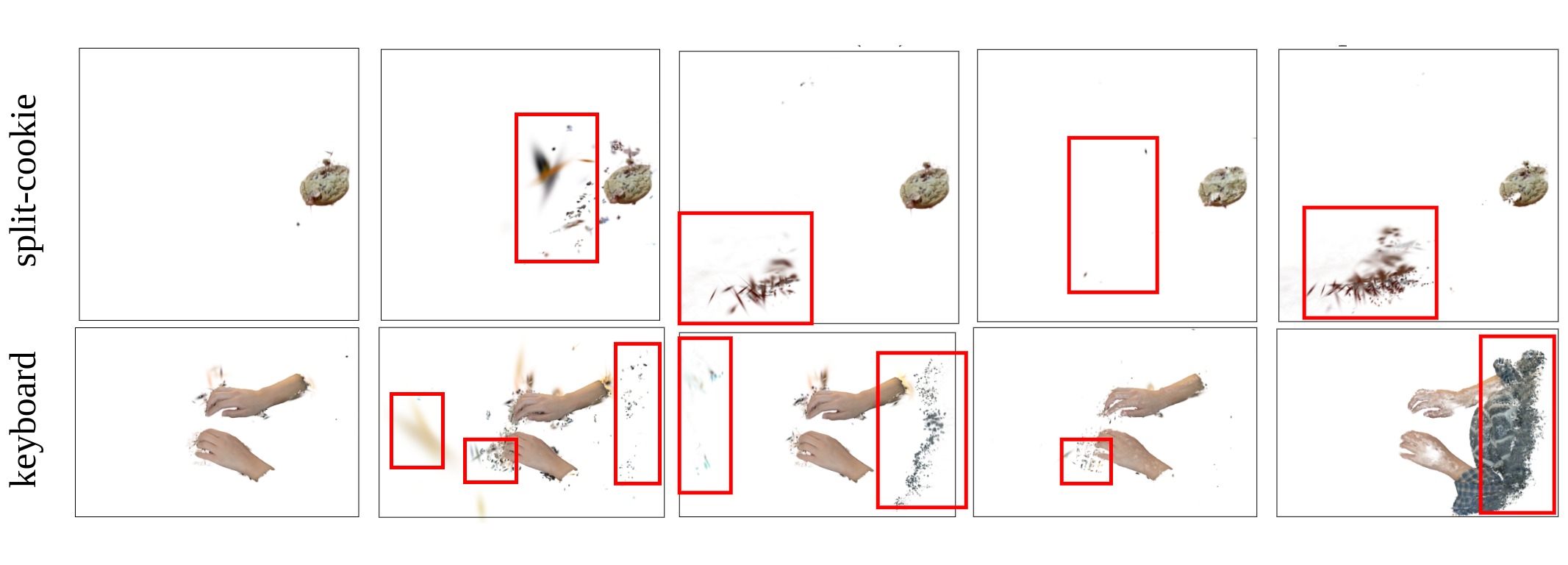}
      \put(6, 35){ {\shortstack{\acronym{} (Ours)}}}
  \put(26, 35){{\shortstack{\acronym{} (hard)}}}
  \put(46, 35){ {\shortstack{\acronym{} (all)}}}
  \put(68, 35){ {\shortstack{CGC}}}
  \put(83, 35){ {\shortstack{OpenGaussian}}}

  \put(0, 18){\rotatebox{90}{{\colorbox{white}{\shortstack{split-cookie}}}}}
  \put(0, 4){\rotatebox{90}{{\colorbox{white}{\shortstack{keyboard}}}}}
  \end{overpic}
  \caption{Additional results of out-of-FoV visualization from \acronym{} with soft sample mining (Ours), hard / all sample mining, CGC \cite{contrastive_gaussians_2024}, and OpenGaussian \cite{wu2025opengaussian}.}
  \label{fig:addtional_out_of_fov}
\end{figure*}

\noindent\textbf{Comparison to Contrastive Methods. }Existing contrastive segmentation methods CGC~\cite{silva2024contrastive} and OpenGaussian \cite{wu2025opengaussian} treat all negative pairs equally and focus on average feature consistency. 
However, this often leads to ambiguous feature assignments (\cref{fig:feature_space_comparison}) and segmentation artifacts (\cref{fig:addtional_out_of_fov}).
In contrast, \acronym{} optimizes per-pixel relationships by introducing soft sample mining as discussed in our main paper.

It prioritizes challenging pixels that have any negative pixel pairs that are close in feature space and any positive pairs that are far apart, leading to more discriminative feature representations. We confirm our findings by evaluating segmentation accuracy in Tab. 1 (main). Our \acronym{} achieves the best segmentation performance across all tested benchmarks, validating the effectiveness of the learned segmentation field. 

\begin{figure*}[ht]
  \centering
  \begin{overpic}[width=0.9\textwidth]{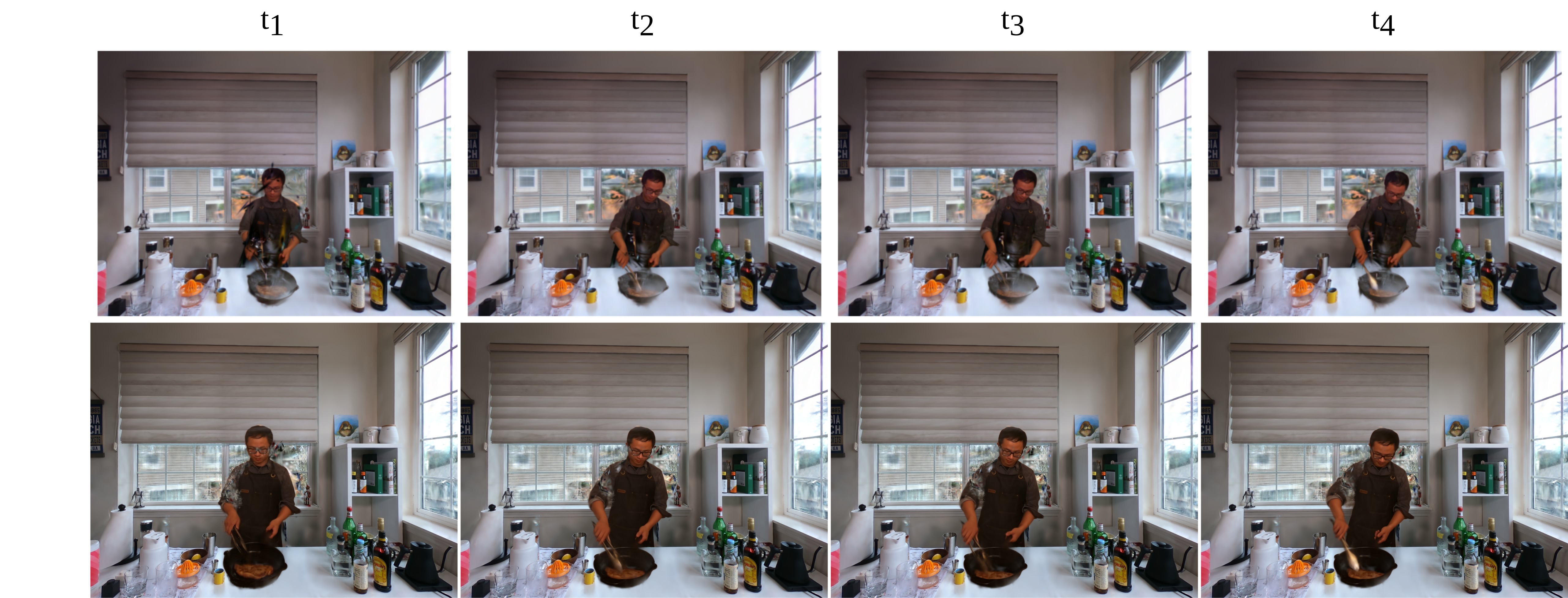}
      \put(0, 21){\rotatebox{90}{{{\shortstack{Composite \\
SA4D}}}}}
\put(0, 2){\rotatebox{90}{{{\shortstack{Composite \\
\acronym{} (Ours)}}}}}
  \end{overpic}
  
  \caption{Scene composition with SA4D \cite{ji2024segment} and our \acronym{} on Neu3D sequences. We segment the person from \textit{sear\_steak} and replace him in \textit{coffee\_martini} with it. Our method delivers fewer artifacts. The scale and position of the object are adapted manually to fit the scene.}
  \vspace{-2mm}
  \label{fig:scene_composition_with_sa4d}
\end{figure*}
\section{Additional Feature Space Visualizations}
\label{sec:supp_feat_space}
\begin{figure*}[t]
  \centering
  \begin{overpic}[width=0.9\textwidth]{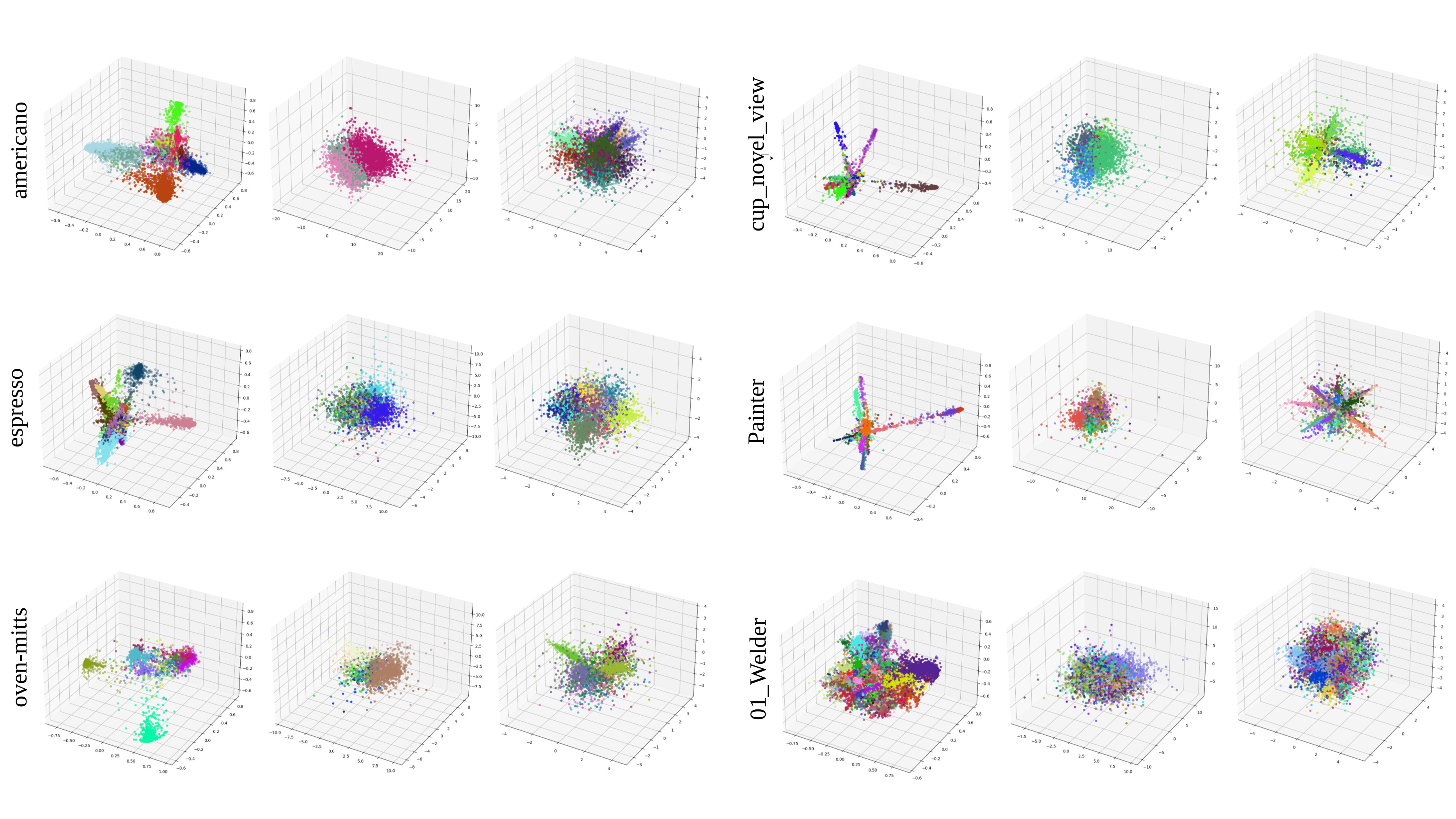}
  \put(7, 54){\scriptsize {\shortstack{\acronym}}}
  \put(23, 54){\scriptsize {\shortstack{CGC}}}
  \put(36, 54){\scriptsize {\shortstack{OpenGaussian}}}
  \put(58, 54){\scriptsize {\shortstack{\acronym}}}
  \put(74, 54){\scriptsize {\shortstack{CGC}}}
  \put(87, 54){\scriptsize {\shortstack{OpenGaussian}}}
  \end{overpic}
  \caption{Additional results of feature space visualization from \acronym{}, CGC \cite{contrastive_gaussians_2024}, and OpenGaussian \cite{wu2025opengaussian}. Our method exhibits more compact clusters with more refined boundaries than other methods, which indicates high feature discrimination of our latent space. }
  \label{fig:addtional_feature_space_visualization}
\end{figure*}

As shown in \cref{fig:feature_space_comparison}, \acronym{} demonstrates superior performance in separating different clusters apart compared to the existing contrastive learning approaches \cite{wu2025opengaussian, contrastive_gaussians_2024}. We provide additional results in \cref{fig:addtional_feature_space_visualization}. The features learned by \acronym{} are more discriminative compared to other approaches. 

\section{Additional Out-of-FoV Visualizations}
\label{sec:supp_outfov}
As shown in the main paper, \acronym{} with soft sample mining demonstrates superior performance in the out-of-FoV viewpoint to eliminate floating artifacts compared to \acronym{} with hard / all sample mining 
and the existing contrastive learning approaches \cite{wu2025opengaussian, contrastive_gaussians_2024}. We provide additional results in \cref{fig:addtional_out_of_fov}.

\begin{table*}[p]
  \centering
  \scalebox{0.8}{
  \begin{tabular}{c c c c c c c c c c c c c}
    \toprule
      & \multicolumn{2}{c}{as\_novel\_view} & \multicolumn{2}{c}{basin\_novel\_view} & \multicolumn{2}{c}{cup\_novel\_view} & \multicolumn{2}{c}{press\_novel\_view} & \multicolumn{2}{c}{plate\_novel\_view} & \multicolumn{2}{c}{Average} \\ 
      Method & mIoU & mAcc & mIoU & mAcc & mIoU & mAcc & mIoU & mAcc & mIoU & mAcc & mIoU & mAcc  \\
    \midrule
      Gaussian Grouping \cite{ye2023gaussian} & 0.8140 & 0.9487 & 0.8083 & 0.9612 & \textbf{0.9020} & \textbf{0.9936} & 0.7803 & 0.9912 & 0.8710 & 0.9696 & 0.8351 & 0.9729\\
      SAGA \cite{cen2023saga} & 0.8235 & 0.9542 & 0.6437 & 0.9140 & 0.8833 & 0.9922 & 0.8119 & 0.9918 & 0.8736 & 0.9699 & 0.8072 & 0.9644\\
      SA4D~\cite{ji2024segment} & 0.7384 &	0.9158	& 0.7663 &	0.9609	& 0.7759 &	0.9858 & 0.6558 & 0.9894 &	0.4337 & 0.8282 & 0.6740 & 0.9360\\
      DGD \cite{labe2024dgd} & 0.7135	& 0.9266 & 0.6685 & 0.9291 & 0.694&0.9752&0.6457&	0.9822&	0.8406&	0.9625&0.7125&	0.9551\\
      CGC~\cite{silva2024contrastive} & 0.7388 &	0.9404	& 0.8084	& 0.9659 &	0.8556 & 0.9908 &	0.7723 &	0.9905 & 0.8748 & 0.9716 & 0.8100 & 0.9718\\
      OpenGaussian~\cite{wu2025opengaussian} & 0.7471 &	0.9438 & 0.8580 & 0.9759 & 0.8406 & 0.9896 &	0.3047 &	0.9500 & 0.7191 & 0.9228 & 0.6939	&0.9564 \\
      \acronym{} (Ours) & \textbf{0.9160} & \textbf{0.9798} & \textbf{0.8704} & \textbf{0.9775} & 0.8893 & 0.9927& \textbf{0.8257} & \textbf{0.9926} & \textbf{0.8826} & \textbf{0.9729} & \textbf{0.8768} & \textbf{0.9831}\\
    \bottomrule
  \end{tabular}
  }
  \caption{Quantitative results for Gaussian Grouping \cite{ye2023gaussian}, SAGA \cite{cen2023saga}, SA4D \cite{ji2024segment}, DGD \cite{labe2024dgd}, CGC \cite{silva2024contrastive}, OpenGaussian~\cite{wu2025opengaussian} and \acronym{} on our \textit{NeRF-DS-Mask} segmentation benchmark. \acronym{} outperforms the baselines on average. }
  \label{tab:nerfds_quantitative}
\end{table*}

\begin{table*}[p]
  \centering
  \scalebox{0.80}{
  \begin{tabular}{c c c c c c c c c c c c c}
    \toprule
      & \multicolumn{2}{c}{americano} & \multicolumn{2}{c}{chickchicken} & \multicolumn{2}{c}{cut-lemon1} & \multicolumn{2}{c}{espresso} & \multicolumn{2}{c}{hand} & \multicolumn{2}{c}{keyboard} \\ 
      Method & mIoU & mAcc & mIoU & mAcc & mIoU & mAcc & mIoU & mAcc & mIoU & mAcc & mIoU & mAcc \\
    \midrule
      Gaussian Grouping \cite{ye2023gaussian} & \textbf{0.8980} & \textbf{0.9969} & \textbf{0.9436} & \textbf{0.9867} & 0.7153 & 0.9394 & 0.4968 & 0.9642 & 0.8865 & 0.9814 & \textbf{0.9159} & \textbf{0.9861}\\
      SAGA \cite{cen2023saga} & 0.7325 & 0.9873 & 0.8633 &0.9648& 0.5263 & 0.8398 & 0.6395 & 0.9798 & \textbf{0.9088} & \textbf{0.9859} & 0.8070 & 0.9645\\
      SA4D~\cite{ji2024segment} & 0.8488 & 0.9941 &	0.8269 &	0.9553 &	0.8161 &	0.9688 &	0.4507 &	0.9640 &	0.7064 &	0.9462 &	0.7930 &	0.9651\\
      DGD \cite{labe2024dgd} & 0.7748	&0.9912&	0.8317&	0.9589&	0.8583&0.9718& 0.7240	& 0.9871 &0.8936&0.9838&0.8451&0.9733\\
      CGC~\cite{silva2024contrastive} & 0.7917 &	0.9915 &	0.8625 &	0.9672 &	0.8047 &	0.9605 &	\textbf{0.7586} &	\textbf{0.9898} &	0.7692 &	0.9646 &	0.8838 &	0.9810\\
      OpenGaussian~\cite{wu2025opengaussian} & 0.5739 &	0.9723 &	0.8309 &	0.9583 &	0.8369 &	0.9659 &	0.5949 &	0.9760 &	0.7687 &	0.9618 &	0.3848 &	0.8051 \\
      \acronym{} (Ours) & 0.8144&	0.9922	&0.9308&	0.9835&\textbf{0.8795}&	\textbf{0.9769}	&0.7164&	0.986&	0.9006&	0.9849&	0.8952&	0.9827\\
       \midrule
       & \multicolumn{2}{c}{oven-mitts} & \multicolumn{2}{c}{slice-banana} & \multicolumn{2}{c}{split-cookie} & \multicolumn{2}{c}{torchocolate} & \multicolumn{2}{c}{Average} & \multicolumn{2}{c}{} \\ 
      Gaussian Grouping \cite{ye2023gaussian} & 0.7420 & 0.9505 & \textbf{0.9095} &  \textbf{0.9721} & \textbf{0.9134} & \textbf{0.9960} &  \textbf{0.9196} & \textbf{0.9981} & 0.8341 & 0.9771 & &\\
      SAGA \cite{cen2023saga} & 0.8667 & 0.9766 & 0.7323 & 0.8960 & 0.8186 & 0.9906 & 0.7652 & 0.9933 & 0.7660 & 0.9579 \\
      SA4D~\cite{ji2024segment} &	0.7257 &	0.9404 &	0.7205 &	0.8983 &	0.8351 &	0.9931 &	0.6474 &	0.9911 & 0.7371	& 0.9616\\
        DGD \cite{labe2024dgd} & 0.8595	&0.9755	&0.8408	&0.9470	&0.8467	&0.9925	&0.8223&	0.9954&0.8297&	0.9777\\
      CGC~\cite{silva2024contrastive} &	0.9025 &	0.9832 &	0.8378 &	0.9469 &	0.7975 &	0.9906 &	0.7488 &	0.9929 & 0.8157	& 0.9768\\
      OpenGaussian~\cite{wu2025opengaussian} &	0.9107 &	0.9845 &	0.7470 &	0.9170 &	0.8816 &	0.9946 &	0.0011 &	0.9440 & 0.6531 &	0.9480\\
        \acronym{} (Ours) & \textbf{0.9295}&	\textbf{0.9876}&	0.8782&	0.9617&	0.8581&	0.9930	&0.8599	&0.9964 & \textbf{0.8663}	&\textbf{0.9845}&  &\\
    \bottomrule
  \end{tabular}
  }
  \caption{Quantitative results for Gaussian Grouping \cite{ye2023gaussian}, SAGA \cite{cen2023saga}, SA4D \cite{ji2024segment}, DGD \cite{labe2024dgd}, CGC \cite{silva2024contrastive}, OpenGaussian~\cite{wu2025opengaussian} and \acronym{} on our \textit{HyperNeRF-Mask} segmentation benchmark. \acronym{} outperforms the baselines on average.}
  \label{tab:hypernerf_quantitative}
\end{table*}

\begin{table*}[p]
  \centering
  \scalebox{0.80}{
  \begin{tabular}{c c c c c c c c c c c c c}
    \toprule
      & \multicolumn{2}{c}{coffee\_martini} & \multicolumn{2}{c}{cook\_spinach} & \multicolumn{2}{c}{cut\_roasted\_beef} & \multicolumn{2}{c}{flame\_steak} & \multicolumn{2}{c}{sear\_steak} & \multicolumn{2}{c}{Average} \\  
      Method & mIoU & mAcc & mIoU & mAcc & mIoU & mAcc & mIoU & mAcc & mIoU & mAcc & mIoU & mAcc \\
    \midrule
    Gaussian Grouping \cite{ye2023gaussian} &  0.8295 & 0.9890 & 0.8974 & 0.9941 & \textbf{0.9512} & \textbf{0.9972} & 0.8279 & 0.9900& \textbf{0.9261} & \textbf{0.9959} & 0.8864 & 0.9932\\
      SAGA \cite{cen2023saga} &  0.2201 & 0.8081 & 0.8125 & 0.9881 & 0.6982 & 0.9767 & 0.8439 & 0.9911& 0.8959 & 0.9941 & 0.6941 & 0.9516\\
      SA4D~\cite{ji2024segment} & 0.8583 &	0.9910 &	0.8987 &	0.9941 &	0.8645 &	0.9914 &	\textbf{0.8898} &	\textbf{0.9940} &	0.9047 &	0.9948 & 0.8832	& 0.9931\\
      DGD \cite{labe2024dgd} &0.7875&	0.9865&	0.8150 &	0.9883&	0.8170 &	0.9877&	0.6771	&0.9776	&0.7638	&0.9854 &0.7721	&0.9851\\
      CGC~\cite{silva2024contrastive} & 0.8359 &	0.9895 &	0.9034 &	0.9945 &	0.9052 &	0.9943 &	0.8530 &	0.9920 &	0.8794 &	0.9934 & 0.8754	& 0.9927\\
      OpenGaussian~\cite{wu2025opengaussian} & 0.8254	&0.9896	&0.6336	&0.9798	&0.9115	&0.9951&	0.8199	&0.9907	&0.8986&	0.9943 & 0.8178	&0.9899\\
      \acronym{} (Ours) & \textbf{0.9120}&	\textbf{0.9948}	&\textbf{0.9129}	&\textbf{0.9951}&	0.9103&	0.9947	&0.8716&	0.9930&	0.9044	&0.9947& \textbf{0.9022}	&\textbf{0.9945}\\
    \bottomrule
  \end{tabular}
  }
  \caption{Quantitative results for Gaussian Grouping \cite{ye2023gaussian}, SAGA \cite{cen2023saga}, SA4D \cite{ji2024segment}, DGD \cite{labe2024dgd}, CGC \cite{silva2024contrastive} and \acronym{} on our \textit{Neu3D-Mask} segmentation benchmark. \acronym{} remains on par with Gaussian Grouping on average and outperforms SAGA by a large margin on average mIoU. }
  \label{tab:neu3d_quantitative}
\end{table*}

\begin{table*}[p]
  \centering
  \scalebox{0.8}{
  \begin{tabular}{c c c c c c c c c c c}
    \toprule
      & \multicolumn{2}{c}{01\_Welder} & \multicolumn{2}{c}{02\_Flames} & \multicolumn{2}{c}{10\_Alexa\_1} & \multicolumn{2}{c}{11\_Alexa\_2} & \multicolumn{2}{c}{Average}\\ 
      Method & mIoU & mAcc & mIoU & mAcc & mIoU & mAcc & mIoU & mAcc & mIoU & mAcc\\
    \midrule
    Gaussian Grouping \cite{ye2023gaussian} & 0.7920 & 0.9736 & 0.8364 & 0.9936 & 0.7391 & 0.9840 & 0.7018 & 0.9590 & 0.7673 & 0.9776\\
      SAGA \cite{cen2023saga} & 0.8062 & 0.9737 & 0.8463 & 0.9940 & 0.4794 & 0.9513 & 0.8260 & 0.9797 & 0.7395 & 0.9747\\
      SA4D~\cite{ji2024segment} & 0.7728 &	0.9708 &	0.7200 &	0.9871 &	0.8351 &	0.9907 &	0.8669 &	0.9853 & 0.7987	& 0.9835\\
      DGD \cite{labe2024dgd} & 0.873&	0.9854&	0.7775&	0.9914	&0.8107&	0.9899&	0.7312&	0.9734& 0.7981	&0.9850\\
      CGC~\cite{silva2024contrastive} & 0.911	& 0.9894 &	0.8520 &	0.9947 &	0.8464 &	0.9924 &	0.9328 &	0.9933 & 0.8856	& 0.9925\\
      OpenGaussian~\cite{wu2025opengaussian} & \textbf{0.9077}	&\textbf{0.9891}	&0.1439	&0.9419	&0.8752&	0.9941&	0.9213&	0.9922&0.7120	&0.9793\\
      \acronym{} (Ours) & 0.8975&	0.9877&	\textbf{0.8808}&	\textbf{0.9955}&	\textbf{0.9367}&	\textbf{0.9970}&	\textbf{0.9786}&	\textbf{0.9979} &\textbf{0.9234}&	\textbf{0.9945}\\
    \bottomrule
  \end{tabular}
  }
  \caption{Quantitative results for Gaussian Grouping \cite{ye2023gaussian}, SAGA \cite{cen2023saga}, SA4D \cite{ji2024segment}, DGD \cite{labe2024dgd}, CGC \cite{silva2024contrastive}, OpenGaussian~\cite{wu2025opengaussian} and \acronym{} on our \textit{Immersive-Mask} segmentation benchmark. \acronym{} outperforms the baselines on average mIoU by a large margin.}
  \label{tab:immersive_quantitative}
\end{table*}

\begin{table*}[p]
  \centering
  \scalebox{0.8}{
  \begin{tabular}{c c c c c c c c c c c}
    \toprule
      & \multicolumn{2}{c}{Birthday} & \multicolumn{2}{c}{Fabien} & \multicolumn{2}{c}{Painter} & \multicolumn{2}{c}{Theater} & \multicolumn{2}{c}{Average} \\ 
      Method & mIoU & mAcc & mIoU & mAcc & mIoU & mAcc & mIoU & mAcc & mIoU & mAcc \\
    \midrule
    Gaussian Grouping \cite{ye2023gaussian} & \textbf{0.9324} & \textbf{0.9908} & 0.9406 & 0.9758 & 0.8759 & 0.9888 & 0.4428 & 0.9646 & 0.7979	& 0.9800\\
      SAGA \cite{cen2023saga} &0.8541 & 0.9779& 0.9371 & \textbf{0.9943} & 0.9670 & 0.9867 & 0.4071 & 0.9579 & 0.7913 & 0.9792\\
      SA4D~\cite{ji2024segment}& 0.8322 &	0.9736 &	0.8709 &	0.9460 &	0.8070 &	0.9808 &	0.7984 &	0.9930 & 0.8271	& 0.9734\\
      DGD \cite{labe2024dgd}&0.7203&	0.9573	&0.9177	&0.9667	&0.7959&	0.9788	&0.6826	&0.9885 &0.7791	&0.9728\\
      CGC~\cite{silva2024contrastive} & 0.7951 & 0.9668 &	0.9163 &	0.9659 &	0.9668 &	0.9971 &	0.8443 &	0.9950 & 0.8806	& 0.9812\\
      OpenGaussian~\cite{wu2025opengaussian} & 0.7008 &	0.9458 &	0.8590 &	0.9438 &	0.8928 &	0.9908 &	\textbf{0.9034} &	\textbf{0.9971} & 0.8390	& 0.9694\\
      \acronym{} (Ours) & 0.8839&	0.9833&	\textbf{0.9734}&	0.9889&	\textbf{0.9719}	&\textbf{0.9976}	&0.8941&	0.9969 &\textbf{0.9308}&	\textbf{0.9917}\\
    \bottomrule
  \end{tabular}
  }
  
  \caption{Quantitative results for Gaussian Grouping \cite{ye2023gaussian}, SAGA \cite{cen2023saga}, SA4D \cite{ji2024segment}, DGD \cite{labe2024dgd}, CGC \cite{silva2024contrastive}, OpenGaussian~\cite{wu2025opengaussian} and \acronym{} on our \textit{Technicolor-Mask} segmentation benchmark. \acronym{} outperforms the baselines on average mIoU by a large margin.}
  \label{tab:technicolor_quantitative}
\end{table*}

\section{Further Scene Editing Examples}
\label{sec:editing}
\begin{figure*}[ht]
  \centering
  \includegraphics[width=0.8\textwidth]{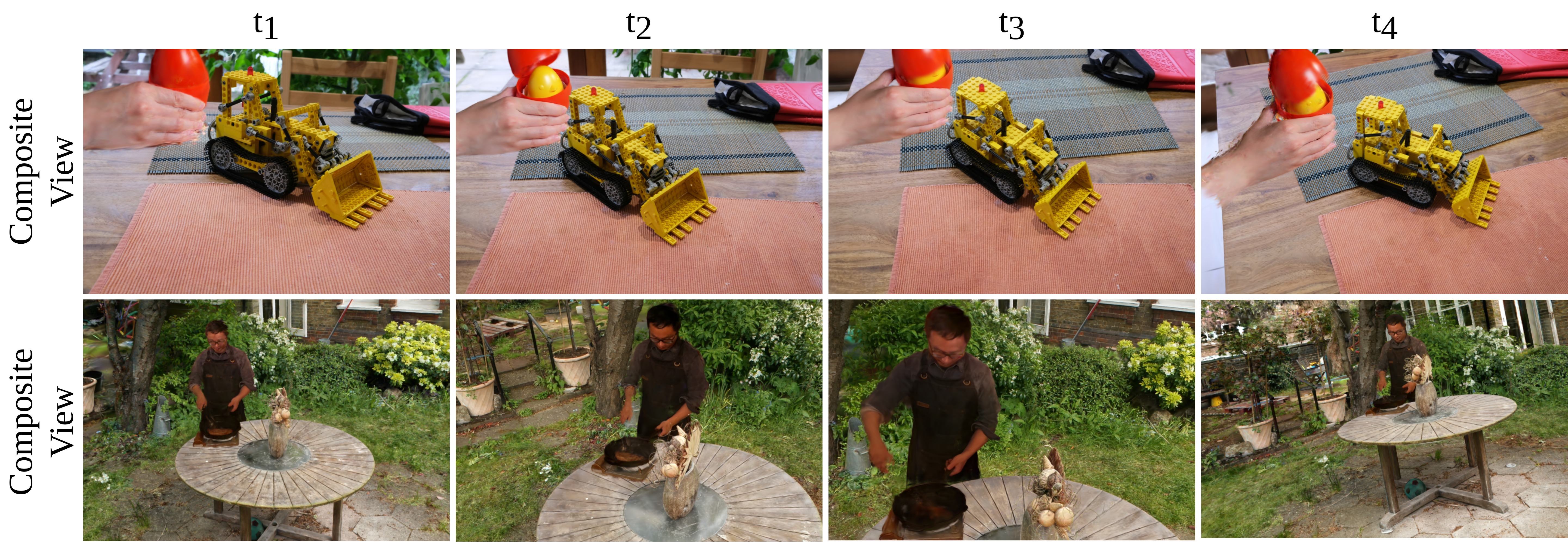}
  \caption{Scene composition with \acronym{}. Top: \textit{chickchicken} and Mip-NeRF 360 \cite{barron2022mipnerf360} \textit{kitchen}. Bottom: \textit{sear\_steak} and Mip-NeRF 360 \cite{barron2022mipnerf360} \textit{garden}. The scale and position of the object are adapted manually to fit the scene.}
  \vspace{-4mm}
  \label{fig:sadg_scene_composition}
\end{figure*}

\begin{figure*}[ht]
  \centering
    \scalebox{0.9}{
    \begin{overpic}[width=0.9\textwidth]{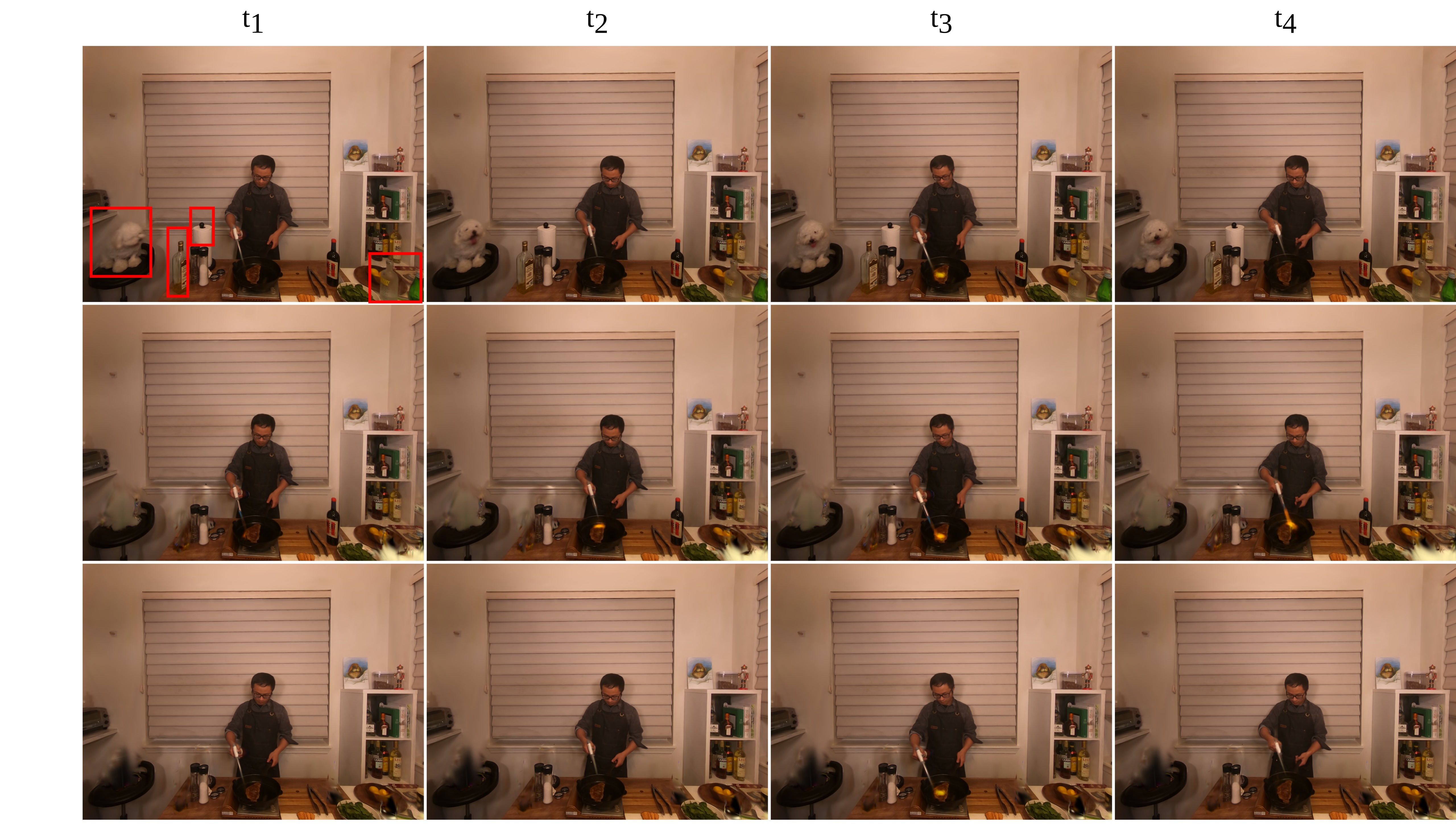}
        \put(0, 41){\rotatebox{90}{{{\shortstack{Rendered \\ View}}}}}
        \put(0, 20){\rotatebox{90}{{{\shortstack{Object Removal \\
SA4D}}}}}
\put(0, 2){\rotatebox{90}{{{\shortstack{Object Removal \\
\acronym{} (Ours)}}}}}
    \end{overpic}
    }
  \caption{Object removal (identified with the red boxes) with SA4D \cite{ji2024segment} and our \acronym{} in Neu3D \textit{flame\_steak} sequence. Our method delivers fewer artifacts and does not introduce ghostly effects in place of the removed object. \vspace{-4mm}}
  \label{fig:object_removal_flame_steak}
\end{figure*}

We present further scene composite examples of \acronym{} in \cref{fig:sadg_scene_composition}. We also compare the quality of object removal and scene composition with SA4D \cite{ji2024segment} in \cref{fig:object_removal_flame_steak} and \cref{fig:object_removal_americano}. While SA4D introduces ghostly artifacts in place of the removed objects and carries spurious Gaussians with the segmented objects, our \acronym{} delivers photorealistic results of the edited scene.

\section{Detailed Quantitative Results per Dataset}
\label{sec:quant_res}
For completeness to the average numbers per datasets (main), we report the performance of our method individually on all benchmarks in~\cref{tab:nerfds_quantitative}, \cref{tab:hypernerf_quantitative}, \cref{tab:neu3d_quantitative}, \cref{tab:immersive_quantitative}, and \cref{tab:technicolor_quantitative}.

\section{Additional Qualitative Results}
\label{sec:further_qualitative}
\begin{figure*}[ht]
  \centering
  \begin{overpic}[width=0.9\textwidth]{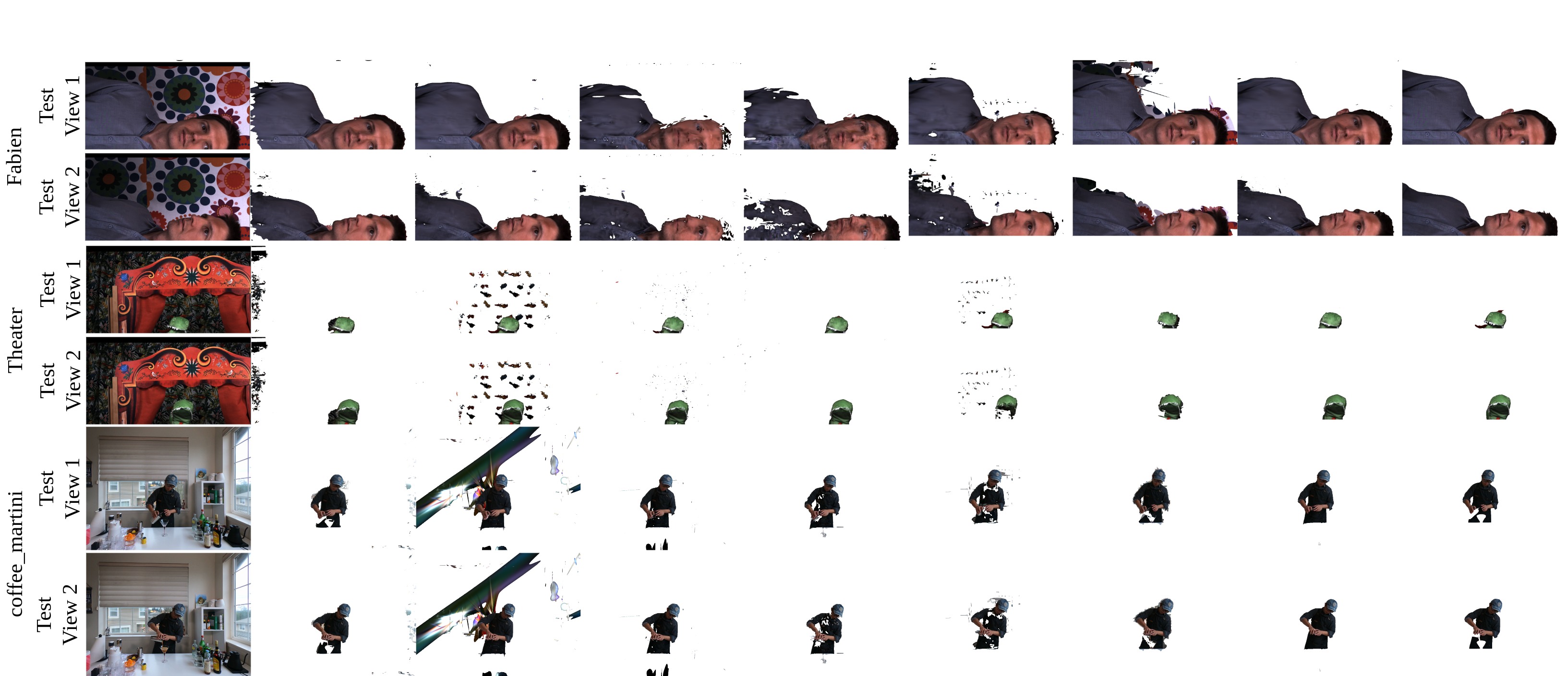}
      \put(7, 41){\scriptsize \colorbox{white}{\shortstack{GT Image}}}
    \put(18, 41){\scriptsize \colorbox{white}{\shortstack{Gaussian \\ Grouping}}}
    \put(29, 41){\scriptsize \colorbox{white}{\shortstack{SAGA}}}
    \put(40, 41){\scriptsize \colorbox{white}{\shortstack{CGC}}}
    \put(47, 41){\scriptsize \colorbox{white}{\shortstack{OpenGaussian}}}
    \put(61, 41){\scriptsize \colorbox{white}{\shortstack{DGD}}}
    \put(71, 41){\scriptsize \colorbox{white}{\shortstack{SA4D}}}
    \put(81, 41){\scriptsize \colorbox{white}{\shortstack{\acronym \\ (Ours)}}}
    \put(91, 41){\scriptsize \colorbox{white}{\shortstack{GT Object}}}
  \end{overpic}
  \caption{Segmentation qualitative results on \textit{Technicolor-Mask} and \textit{Neu3D-Mask} sequences. Our method delivers the best temporally consistent performance across all baselines. Please note that OpenGaussian is not evaluated on \textit{Neu3D-Mask}.}
  \label{fig:additional_qualitative_1}
\end{figure*}

\begin{figure*}[t]
  \centering
  \begin{overpic}[width=0.9\textwidth]{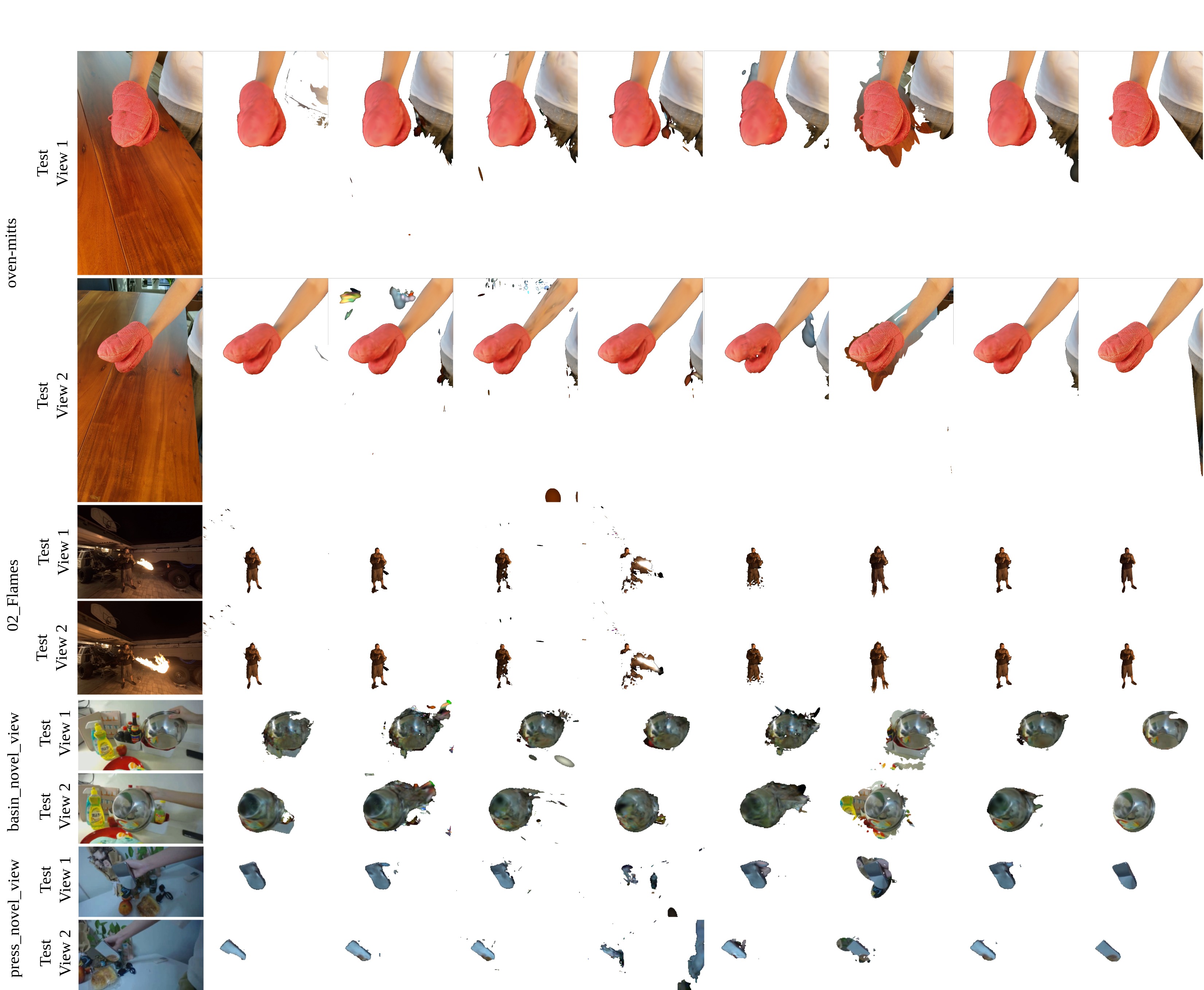}
      \put(8, 80){\scriptsize \colorbox{white}{\shortstack{GT Image}}}
    \put(19, 80){\scriptsize \colorbox{white}{\shortstack{Gaussian \\ Grouping}}}
    \put(30, 80){\scriptsize \colorbox{white}{\shortstack{SAGA}}}
    \put(40, 80){\scriptsize \colorbox{white}{\shortstack{CGC}}}
    \put(48, 80){\scriptsize \colorbox{white}{\shortstack{OpenGaussian}}}
    \put(61, 80){\scriptsize \colorbox{white}{\shortstack{DGD}}}
    \put(71, 80){\scriptsize \colorbox{white}{\shortstack{SA4D}}}
    \put(81, 80){\scriptsize \colorbox{white}{\shortstack{\acronym \\ (Ours)}}}
    \put(91, 80){\scriptsize \colorbox{white}{\shortstack{GT Object}}}
  \end{overpic}
  \caption{Segmentation qualitative results on \textit{HyperNeRF-Mask}, \textit{Immersive-Mask}, and \textit{NeRF-DS-Mask} datasets. Our method delivers the best temporally consistent performance across all baselines.}
  \label{fig:additional_qualitative_2}
\end{figure*}

\begin{figure*}[t]
  \centering
  \begin{overpic}[width=0.9\textwidth]{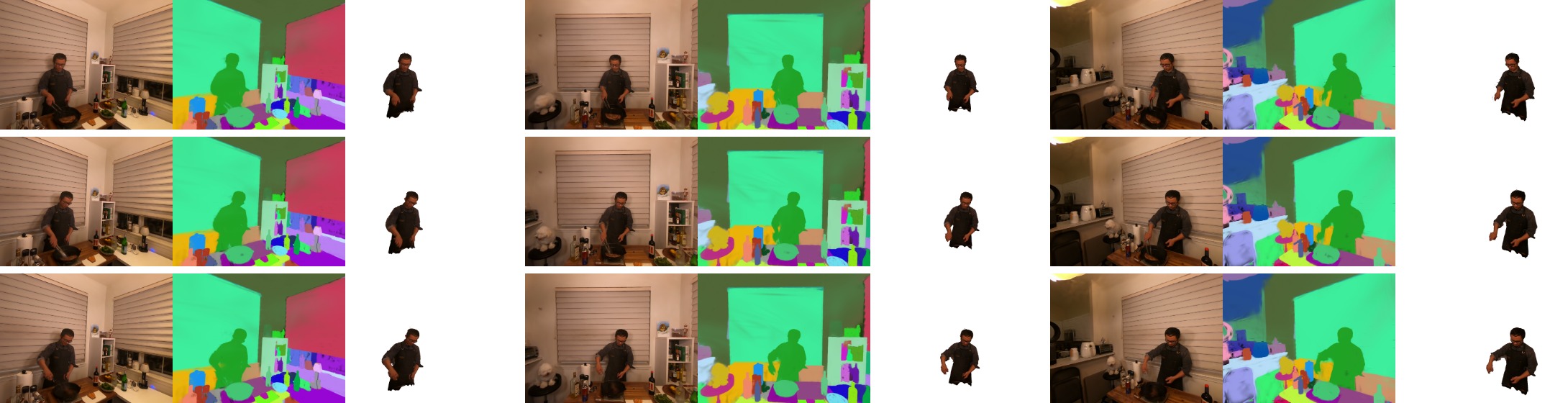}
  \put(-3, 19){\rotatebox{90}{$t = 0$}}
  \put(-3, 9.5){\rotatebox{90}{$t = 0.5$}}
  \put(-3, 2){\rotatebox{90}{$t = 1$}}
  \put(11, 27){Left Camera}
  \put(43, 27){Center Camera}
  \put(77, 27){Right Camera}
  \end{overpic}
  \caption{Visualization of the rendered scene from the left, center, and right camera. For each camera, we show the rendered view (left), rendered segmentation (middle), and segmented object (right). \acronym{} demonstrate multiview consistency in time and space. }
  \label{fig:multi_view_concat}
\end{figure*}

We provide additional qualitative results in \cref{fig:additional_qualitative_1} and \cref{fig:additional_qualitative_2} to further showcase the effectiveness of \acronym{} over other methods. While SAGA~\cite{cen2023segment} suffers from degenerate Gaussians, our reconstruction backbone and effective post-processing mitigate the issue. At the same time, rendered masks from other methods, such as DGD~\cite{labe2024dgd} and SA4D~\cite{ji2024segment}, are inaccurate and introduce spurious regions. Our method demonstrates the best performance among all approaches. We also showcase in \cref{fig:multi_view_concat} that \acronym{} achieves both time and space consistency in multiview video dataset such as Neu3D. 

\begin{figure*}[ht]
  \centering
  \begin{overpic}[width=0.8\textwidth]{fig/object_removal_americano.jpg}
      \put(2, 78){\rotatebox{90}{{{\shortstack{Rendered \\ View}}}}}
        \put(2, 44){\rotatebox{90}{{{\shortstack{Object Removal \\
SA4D}}}}}
\put(2, 10){\rotatebox{90}{{{\shortstack{Object Removal \\
\acronym{} (Ours)}}}}}
  \end{overpic}
  \caption{Object removal (saucer in the red box) with SA4D \cite{ji2024segment} and our \acronym{} in HyperNeRF \textit{americano} sequence. Our method delivers fewer artifacts and does not introduce ghostly effects in place of the removed object. Our method does not require additional inpainting as the background is complete due to multi-view observations. }
  \label{fig:object_removal_americano}
\end{figure*}

\section{Visualization of the Rendered View, Rendered Segmentation, Segmented Object}
\label{sec:visualization_gaussian_clusters}
\begin{figure*}[ht]
  \centering
  \includegraphics[width=0.9\textwidth]{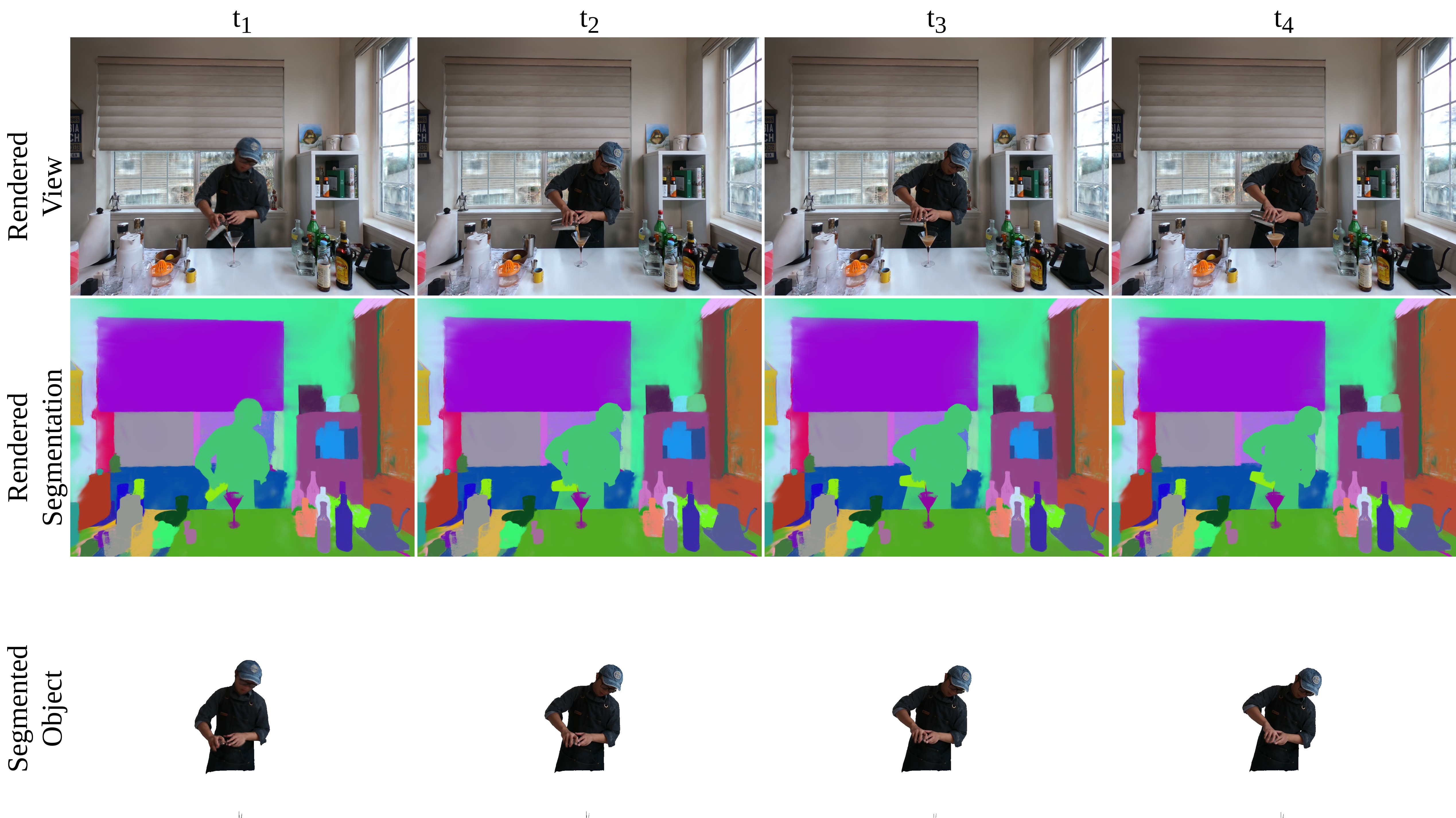}
  \caption{Visualization of rendered view, rendered segmentation, and segmented objects of the scene \textit{coffee\_martini} in four different timestamps in the novel views. The consistency of the semantic feature field across time opens out-of-box editing opportunities for dynamic scenes. }
  \label{fig:visualization1}
\end{figure*}

\begin{figure*}[t]
  \centering
  \includegraphics[width=0.9\textwidth]{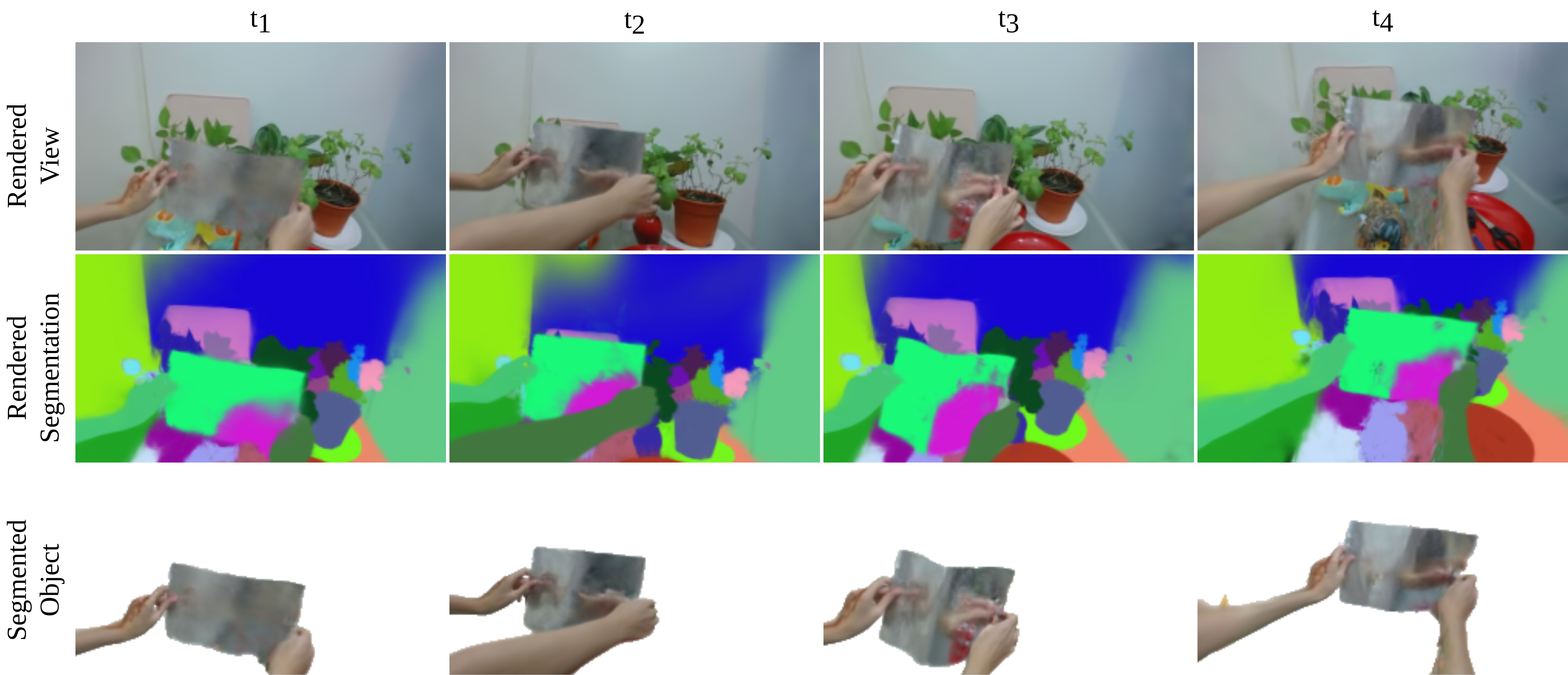}
  \caption{Visualization of rendered view, rendered segmentation, and segmented objects of the scene \textit{as\_novel\_view} in four different timestamps in the novel views. The reflection in the foil remains temporally consistent, which enables post-processing editing effects.}
  \label{fig:visualization4}
\end{figure*}

\begin{figure*}[t]
  \centering
  \includegraphics[width=0.9\textwidth]{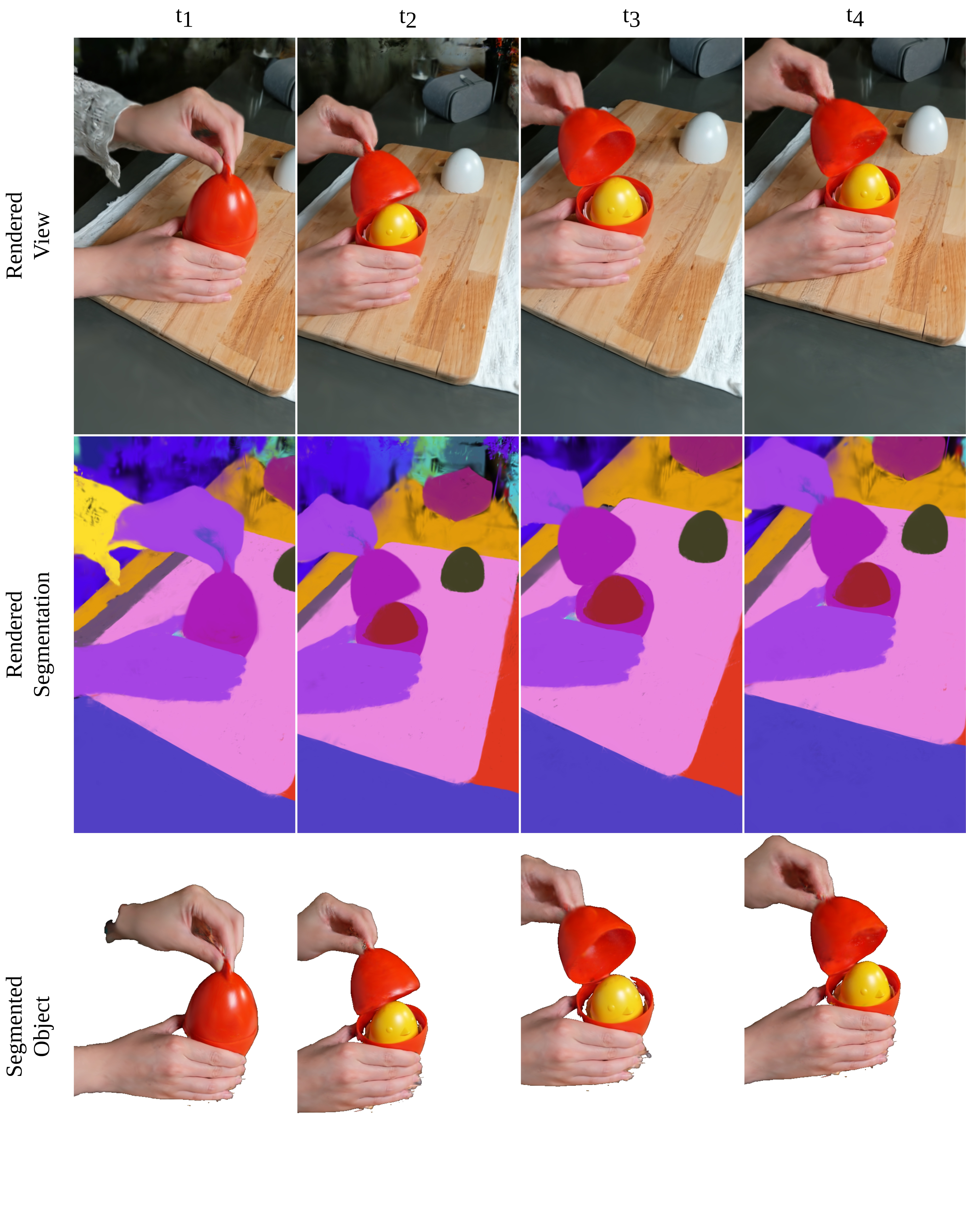}
  \caption{Visualization of rendered view, rendered segmentation, and segmented objects of the scene \textit{chickchicken} in four different timestamps in the novel views. Despite severe occlusions, semantic segmentation is consistent over time.}
  \label{fig:visualization5}
\end{figure*}

We also provide the visualizations of the rendered novel view, the rendered segmentation, and the segmented objects in \cref{fig:visualization1}, \cref{fig:visualization4}, and \cref{fig:visualization5}. The rendered segmentation of \acronym{} is based on Gaussian clustering, which is performed in 3D space, resulting in consistent segmentation in the dynamic scene. 

